\documentclass[11pt]{article}
\usepackage[letterpaper,margin=1in]{geometry}
\usepackage[T1]{fontenc}
\usepackage[utf8]{inputenc}
\usepackage{lmodern,microtype,setspace}
\usepackage{amsmath,amssymb,mathtools,amsfonts,amsthm}
\usepackage{booktabs,array,longtable,tabularx,pdflscape,graphicx,caption}
\usepackage{algorithm,algpseudocode,enumitem,flafter,needspace}
\usepackage[section]{placeins}
\usepackage[square,numbers,sort&compress]{natbib}
\usepackage[hidelinks]{hyperref}
\usepackage[nameinlink,capitalise]{cleveref}
\theoremstyle{plain}

\newtheorem{proposition}{Proposition}

\theoremstyle{remark}

\crefname{assumption}{Assumption}{Assumptions}
\Crefname{assumption}{Assumption}{Assumptions}
\DeclareMathOperator{\tr}{tr}
\DeclareMathOperator{\rank}{rank}

\DeclareMathOperator{\sym}{sym}
\DeclareMathOperator{\qf}{qf}

\newcommand{\R}{\mathbb{R}}
\newcommand{\E}{\mathbb{E}}
\newcommand{\St}{\mathrm{St}}

\newcommand{\opnorm}[1]{\left\|#1\right\|_{\mathrm{op}}}

\providecommand{\mainref}[1]{\cref{#1}}
\providecommand{\suppref}[1]{\cref{#1}}
\newcommand{\PaperTitle}{Online Supervised Dimension Reduction with Random Features: Diagnostics and Computational Trade-offs}
\author{Zhenlin Yao, Wei Xiong\\School of Statistics\\University of International Business and Economics\\Beijing, China}
\date{}
\hypersetup{pdftitle={Online Supervised Dimension Reduction with Random Features: Diagnostics and Computational Trade-offs},pdfauthor={Zhenlin Yao, Wei Xiong},pdfsubject={Same-target diagnostics and conditional numerical basis access},pdfkeywords={supervised dimension reduction, random features, cross-covariance, subspace tracking},pdfdisplaydoctitle=true}
\newcommand{\StartSupplement}{%
  \setcounter{section}{0}\setcounter{equation}{0}\setcounter{figure}{0}\setcounter{table}{0}%
  \renewcommand{\thesection}{S\arabic{section}}%
  \renewcommand{\theequation}{S\arabic{equation}}%
  \renewcommand{\thefigure}{S\arabic{figure}}%
  \renewcommand{\thetable}{S\arabic{table}}%
  \renewcommand{\theHsection}{supp.\arabic{section}}%
  \renewcommand{\theHequation}{supp.\arabic{equation}}%
  \renewcommand{\theHfigure}{supp.\arabic{figure}}%
  \renewcommand{\theHtable}{supp.\arabic{table}}%
}

\title{\PaperTitle}
\begin{document}
\maketitle
\begin{abstract}
Accurate optimization of a supervised spectral objective need not produce an accurate population subspace or a better predictive representation. We investigate these distinctions for Online Kernel Supervised Principal Component Analysis (OKSPCA), which combines a centered cross-moment in finite random-feature coordinates with an Adam-style orthonormal basis update for an established objective. Fixed-map consistency, concentration and perturbation results describe the estimator and its exact subspace; same-target comparisons then assess the practical iterate separately. Across six predictive benchmarks, performance depends on the declared pipeline: replacing the tracker with the exact empirical target leaves the two regression deficits largely unchanged. Direct classification-rank models capture nearly all terminal objective energy on average, but a saved intermediate state exhibits substantial geometric deviation; a controlled sample-size study further separates empirical accuracy from population recovery. In distinct numerical-service workloads, exact on-request computation is faster in the tested classification settings, whereas Adam saves time relative to the tested full thin-SVD service for some dense wider-regression requests, alongside persistent geometric error. These diagnostics limit explanations based solely on terminal optimization accuracy and distinguish numerical cost from quality, rank coverage and freshness; they establish neither practical-tracker convergence nor predictive or deployment benefits from basis availability.

\end{abstract}
\section{Introduction}\label{sec:introduction}
\subsection{Motivation}
Supervised dimension reduction seeks predictor coordinates that preserve response-associated information. Supervised PCA, HSIC-based supervised embeddings and covariance-based methods offer several ways to express this association \citep{Bair2006SPC,Barshan2011SPCA,PapazoglouYin2025CSPCA,GhoshKirby2022Centroid}. In a stream, explicit features permit the associated empirical moments to be updated without retaining a growing sample-level Gram matrix. An orthonormal basis can then be maintained approximately, computed when requested, or cached between factorizations. These choices raise two separate questions: how accurately the basis solves its empirical objective, and whether improving that solution changes the prediction or numerical service for which it is used.

This paper studies those questions through a common empirical target. A same-target comparison holds the accumulated moment fixed and replaces the maintained subspace with its exact spectral solution. It can therefore test whether a particular discrepancy is explained by terminal tracking accuracy, without treating different supervised criteria as interchangeable. Prediction remains a property of the complete representation and evaluation pipeline. Likewise, a fast basis response can be stale, rank-limited, or geometrically inaccurate. The purpose is to determine which explanations the observed comparisons support, rather than to infer usefulness from objective capture alone.

\subsection{Related work}
The operator is established. In finite coordinates, HSIC-supervised PCA has a centered cross-product form \citep{Barshan2011SPCA,gretton2005}; SRP and KSRP also use centered explicit-feature products for supervised embeddings \citep{Karimi2018SRP}. The positive-spectrum input subspace agrees with that of orthonormal covariance-maximizing two-block PLS, although the respective objectives sum squared singular values and singular values. CCA additionally whitens marginal covariances \citep{Hotelling1936CCA,FrankFriedman1993Chemometrics,Rosipal2001KPLS}. Random Fourier features provide the nonlinear coordinates, not a new supervised objective \citep{Rahimi2007RFF}.

Sequential spectral computation also has substantial precedents: recursive and incremental PLS \citep{DayalMacGregor1997REWPLS,Qin1998RPLS,wang2003recursive,Zeng2014IPLS}, stochastic PLS \citep{arora2012stochastic,Arora2016SAPLS,Chen2017OnlinePLS}, paired-stream singular-subspace tracking \citep{Kaiser2010CoupledSVD,Feng2017PSS}, online supervised reduction \citep{Xie2015OSDR}, and online SIR and kernel SIR \citep{Cai2020OnlineSIR,Xu2025OnlineKSIR}. Incremental SVD, online PCA and streaming kernel PCA supply related numerical mechanisms \citep{brand2006fast,Oja1982PCA,Chin2007IKPCA,Ullah2018StreamingKPCA,Deng2025StreamingKPCA}. We use covariance-free incremental PLS (CIPLS) as a predictive comparator \citep{Jordao2021CIPLS}; it is not an exhaustive representation of these alternatives. Existing KSPCA population-objective bounds likewise differ from a guarantee for the practical tracker considered here \citep{Ashtiani2015KSPCABound}.

\subsection{Study objectives and contributions}
OKSPCA combines recursively centered moments with a Stiefel-constrained Adam-style update. We first define this actual computation and state its fixed-map statistical foundation (\cref{sec:method}). Three comparisons then organize the evidence. First, can exact solution of the same terminal moment explain the predictive gaps in the declared pipelines? The two regression deficits largely survive this replacement. Second, do high terminal objectives describe the maintained geometry? Direct classification-rank reconstructions include a marked intermediate deviation, while a controlled sample-size study separates empirical and population subspaces (\cref{sec:prediction,sec:population}). Third, what is the cost of supplying a basis under common request schedules? Exact on-request, projected gradient, periodic caches and applicable structured exact services expose conditional trade-offs that differ between the numerical workloads (\cref{sec:services}).

The contribution is this set of controlled empirical-target diagnostics and its computational consequences, not a new HSIC/PLS objective or a general superiority claim. The predictive comparisons retain their adapter, coordinatewise evaluator and tuning differences. The service study varies several workload factors jointly and compares specified implementations, including full thin SVD; it does not isolate target width or establish competitiveness against all leading-$k$ spectral solvers. These distinctions determine how far the findings can be generalized.

\section{Methodology and theoretical properties}\label{sec:method}

\subsection{Supervised cross-covariance objective}

The diagnostic comparisons require an explicit distinction between the operator being estimated, an exact basis for that operator, and the basis returned by an implementation. We first define their common objective, then give the causal update actually used in the experiments. The final two subsections explain what objective capture can say about geometry and which statistical guarantees require a different, fixed-map setting. Prediction is evaluated separately because an objective on cross-moments does not specify a predictive loss or its fitted evaluator.

Let $\phi(x)\in\R^{D_x}$ and $\psi(y)\in\R^{D_y}$ be finite feature coordinates. With fixed maps and preprocessing, define
\begin{equation}\label{eq:population_cross_moment}
C_*=\E[(\phi-\mu_\phi)(\psi-\mu_\psi)^\top],\qquad M_*=C_*C_*^\top.
\end{equation}
For $1\le k\le D_x$, the established supervised spectral objective is
\begin{equation}\label{eq:population_objective}
\max_{U\in\St(D_x,k)}\|C_*^\top U\|_F^2,
\qquad \St(D_x,k)=\{U:U^\top U=I_k\}.
\end{equation}
An exact maximizer spans leading left singular directions, with possible nonuniqueness at a tied boundary. For any moment $C$ and $1\le q\le\rank(C)$, orthonormal covariance-maximizing two-block PLS selects the same input-side maximizing subspaces: its criterion sums the leading singular values, whereas \cref{eq:population_objective} sums their squares \citep{fan1949,overton1992sum,arora2012stochastic,Arora2016SAPLS}. This relation concerns the optimization problems, not arbitrary PLS/CIPLS implementations or trajectories; CCA additionally whitens marginal covariances.

For feature-column matrices $Z,W$, $H=I_n-\mathbf1\mathbf1^\top/n$ and $L=W^\top W$, the Bessel-normalized empirical moment satisfies
\begin{equation}\label{eq:finite_hsic_operator}
C_n=(n-1)^{-1}ZHW^\top,\qquad
C_nC_n^\top=(n-1)^{-2}ZHLHZ^\top\quad(n\ge2).
\end{equation}
This is the finite-coordinate HSIC supervised-PCA operator \citep{Barshan2011SPCA}. Random-feature supervised embeddings, including SRP/KSRP centered cross-products, precede the present construction \citep{Karimi2018SRP}. Our comparisons concern this common finite-feature target; neither the objective nor the centering identity is claimed as new. The analysis supplies no full-kernel approximation guarantee.

The cross-product used by a direct supervised embedding need not itself be an orthonormal leading-subspace basis. Conversely, agreement of optimal input-side subspaces does not require two methods to return identical coordinate systems. Scaling, rotation, feature construction and the downstream fitting rule can therefore remain relevant even when two methods are related through the same spectral operator. In OKSPCA, ``kernel'' denotes the stationary kernels approximated by the sampled random Fourier features (RFF), whose finite coordinates remain the evaluation target.

Classification uses unweighted one-hot labels, $D_y=C$. For positive class probabilities $p_c$, feature means $\mu_c$ within classes and overall mean $\mu$,
\begin{equation}\label{eq:class_mean_weights}
C_*[:,c]=p_c(\mu_c-\mu),\qquad
M_*=\sum_c p_c^2(\mu_c-\mu)(\mu_c-\mu)^\top.
\end{equation}
Thus conditional feature-mean contrasts receive squared-probability weights, not the weights of ordinary between-class scatter. Zero-probability classes contribute zero columns. This criterion guarantees neither sufficient prediction nor macro-F1 optimality. Since centered labels sum to zero,
\begin{equation}\label{eq:onehot_nullspace}
C_*\mathbf1_C=0,\qquad \bar C_t\mathbf1_C=0,\qquad
\rank(C_*),\rank(\bar C_t)\le C-1.
\end{equation}
The sample identity is pathwise even with evolving input preprocessing. The cap $C-1$ is an upper bound, not an asserted rank or positive gap. If $k>r=\rank(C_*)$, exact maximizers include a generally nonunique $(k-r)$-dimensional null-space completion; at $k=D_x$ the whole space is fixed. Rotated basis columns cannot individually be designated as supervised or completion coordinates.

\subsection{Online moment estimation and basis updates}

At observation $t$, input and regression-target standardizers $S^x_{t-1},S^y_{t-1}$ use only prior observations. The input map is
\begin{equation}\label{eq:implemented_feature_maps}
\phi_t=\sqrt{2/D_x}\cos(W_xS^x_{t-1}(x_t)+b_x).
\end{equation}
The entries of $W_x\in\R^{D_x\times d_x}$ are sampled from $N(0,\sigma_x^{-2})$, and phases from $\mathrm{Unif}(0,2\pi)$, once per run. Regression uses the analogous map of its causally standardized scalar response; classification uses its declared one-hot vector. Starting from zero means and raw moment, update
\begin{equation}\label{eq:running_moments}
\begin{aligned}
\mu_{\phi,t}&=(1-t^{-1})\mu_{\phi,t-1}+t^{-1}\phi_t,\\
\mu_{\psi,t}&=(1-t^{-1})\mu_{\psi,t-1}+t^{-1}\psi_t,\\
R_t&=(1-t^{-1})R_{t-1}+t^{-1}\phi_t\psi_t^\top,
&\bar C_t&=R_t-\mu_{\phi,t}\mu_{\psi,t}^\top.
\end{aligned}
\end{equation}
Induction and expansion give
\begin{equation}\label{eq:pathwise_centered_moment}
\bar C_t=t^{-1}\sum_{i=1}^t(\phi_i-\mu_{\phi,t})(\psi_i-\mu_{\psi,t})^\top.
\end{equation}
This exact identity retains the history of realized features. It is not the covariance obtained by replaying every observation through the final standardizer. Set $C_1=0$ and $C_t=t\bar C_t/(t-1)$ thereafter. The positive Bessel rescaling preserves singular subspaces and normalized objective ratios, but is an unbiasedness correction only under the fixed-map assumptions below.

The random weights are fixed, but raw-to-feature coordinates evolve with the standardizer. Prior-only preprocessing is therefore part of the operator's definition: exact factorization removes subspace-optimization error, not the effects of its feature history.

For the current $C_t$, the Frobenius gradient is $2C_t(C_t^\top U)$. With $\sym(A)=(A+A^\top)/2$, its tangent projection and subsequent global clipping are
\begin{equation}\label{eq:practical_gradient}
\begin{aligned}
G_t&=2C_t(C_t^\top U_{t-1}),\\
G_{T,t}&=G_t-U_{t-1}\sym(U_{t-1}^\top G_t),\\
\widetilde G_{T,t}&=
\begin{cases}G_{T,t},&\|G_{T,t}\|_F\le c,\\
cG_{T,t}/(\|G_{T,t}\|_F+10^{-12}),&\|G_{T,t}\|_F>c.
\end{cases}
\end{aligned}
\end{equation}
The optimizer-call counter advances to $s=s_t$ before bias correction. With elementwise operations, the actual update is
\begin{equation}\label{eq:adam_displacement}
\begin{aligned}
m_s&=\beta_1m_{s-1}+(1-\beta_1)\widetilde G_{T,t},\\
v_s&=\beta_2v_{s-1}+(1-\beta_2)\widetilde G_{T,t}^{\odot2},\\
\widehat m_s&=m_s/(1-\beta_1^s),&\widehat v_s&=v_s/(1-\beta_2^s),\\
\Delta_t&=\alpha\widehat m_s\oslash(\sqrt{\widehat v_s}+\epsilon),
&U_t&=\qf_+(U_{t-1}+\Delta_t).
\end{aligned}
\end{equation}
Signed reduced QR, denoted $\qf_+$, makes the diagonal of its triangular factor nonnegative. The arrays remain in ambient coordinates without transport. Elementwise preconditioning need not preserve tangency or right-rotation equivariance, and a finite step need not increase the current objective. We therefore call this a Stiefel-constrained Adam-style update, not a Riemannian-Adam convergence construction \citep{Absil2008Manifolds,kingma2014adam,Becigneul2018riemannian}. The associative gradient multiplication uses the current plug-in moment; it is not a fresh unbiased observation of the population gradient. In particular, the quadratic operation $C\mapsto CC^\top$ does not preserve finite-sample unbiasedness of $C$.

For \cref{alg:okspca_practical}, write $\phi_x(z)=\sqrt{2/D_x}\cos(W_xz+b_x)$ and $\psi_y(z)=\sqrt{2/D_y}\cos(W_yz+b_y)$ for the fixed RFF maps, and let $\kappa(y)$ be the declared class index. The operator $\operatorname{Clip}_c$ is the piecewise map in \cref{eq:practical_gradient}. In the Adam steps, squares, square roots and $\oslash$ act elementwise. The operation $\operatorname{WF}$ is a Welford state update with the small-count, variance-divisor and threshold conventions in \suppref{supp:proofs}; it does not transform the current observation again.

\begin{algorithm}[tbp]
\caption{One-observation OKSPCA update: normal finite-data path}
\label{alg:okspca_practical}
\begin{algorithmic}[1]
\Require $(x_t,y_t)$, index $t$; fixed map parameters and class encoding; $\alpha,c,\beta_1,\beta_2,\epsilon$.
\Statex \textbf{State:} $U_{t-1},\mu_{\phi,t-1},\mu_{\psi,t-1},R_{t-1},m_{s_{t-1}},v_{s_{t-1}},s_{t-1},S^x_{t-1}$; also $S^y_{t-1}$ for regression.
\Statex \textbf{Initialization:} $U_0=\qf_+(Z_0)$ for seeded Gaussian $Z_0\in\R^{D_x\times k}$; feature means, raw moment, Adam arrays and counters zero; raw standardizers empty.
\Statex \textbf{A. Form features using prior standardizers}
\State $\phi_t\gets\phi_x(S^x_{t-1}(x_t))$
\State $\displaystyle\psi_t\gets\begin{cases}e_{\kappa(y_t)},&\text{classification},\\\psi_y(S^y_{t-1}(y_t)),&\text{regression}.
\end{cases}$
\Statex \textbf{B. Accumulate the realized-feature cross-moment}
\State $\mu_{\phi,t}\gets(1-t^{-1})\mu_{\phi,t-1}+t^{-1}\phi_t,\quad
\mu_{\psi,t}\gets(1-t^{-1})\mu_{\psi,t-1}+t^{-1}\psi_t$
\State $R_t\gets(1-t^{-1})R_{t-1}+t^{-1}\phi_t\psi_t^\top$
\State $\displaystyle\bar C_t\gets R_t-\mu_{\phi,t}\mu_{\psi,t}^\top,\quad
C_t\gets\begin{cases}0,&t=1,\\t\bar C_t/(t-1),&t\ge2.\end{cases}$
\Statex \textbf{C. Maintain the basis with ambient Adam}
\State $G_t\gets2C_t(C_t^\top U_{t-1})$
\State $G_{T,t}\gets G_t-U_{t-1}\sym(U_{t-1}^\top G_t)$
\State $\widetilde G_{T,t}\gets\operatorname{Clip}_c(G_{T,t}),\quad s_t\gets s_{t-1}+1$
\State $m_{s_t}\gets\beta_1m_{s_{t-1}}+(1-\beta_1)\widetilde G_{T,t}$
\State $v_{s_t}\gets\beta_2v_{s_{t-1}}+(1-\beta_2)\widetilde G_{T,t}^{\odot2}$
\State $\displaystyle\widehat m_{s_t}\gets\frac{m_{s_t}}{1-\beta_1^{s_t}},\quad
\widehat v_{s_t}\gets\frac{v_{s_t}}{1-\beta_2^{s_t}}$
\State $\Delta_t\gets\alpha\widehat m_{s_t}\oslash(\sqrt{\widehat v_{s_t}}+\epsilon),\quad
U_t\gets\qf_+(U_{t-1}+\Delta_t)$
\Statex \textbf{D. Advance raw standardizers last}
\State $S^x_t\gets\operatorname{WF}(S^x_{t-1},x_t)$
\State $S^y_t\gets\operatorname{WF}(S^y_{t-1},y_t)$ for regression only.
\Ensure Updated carried state; $U_t$ is the maintained basis, not necessarily a later query's returned block.
\end{algorithmic}
\end{algorithm}

\begin{samepage}
On this normal path, $s_t=t$; the optimizer counter is nevertheless a distinct state variable. The first zero-gradient observation advances it and executes signed QR, whose zero diagonal receives sign $+1$. Settings and numerical-failure branches remain in \suppref{supp:proofs}; query-time extraction and checks are separate service operations. Evolving scalers, plug-in moments, clipping, fixed learning rate and ambient adaptive arrays fall outside the standard stochastic-retraction convergence assumptions; convergence of this coupled practical recursion is not established \citep{bonnabel2013sgd,Vary2024RandomStiefel,SakaiIiduka2025Riemannian}.
\par\end{samepage}

\subsection{Objective values and subspace geometry}

For $U\in\St(D_x,q)$, define
\begin{equation}\label{eq:normalized_objective}
J_q(U;C)=\|C^\top U\|_F^2,\qquad J_q^*(C)=\sum_{j=1}^q\sigma_j(C)^2,
\qquad \rho_q=J_q/J_q^*.
\end{equation}
Singular spectra are extended by zeros; the ratio requires $J_q^*>0$. An exact reference is a leading subspace of the same moment. To extract the best $q\le k$ block within a maintained $U\in\St(D_x,k)$, put $B_U=U^\top\bar C_t$. The Ky Fan principle gives
\begin{equation}\label{eq:within_span_variational}
\max_{Q\in\St(k,q)}J_q(UQ;\bar C_t)=\sum_{j=1}^q\sigma_j(B_U)^2.
\end{equation}
Leading left singular vectors of $B_U$ give the coefficient block, followed by signed QR of $UQ$. The value uses the projected spectrum, not the unrestricted moment spectrum; uniqueness can fail even at a zero tie. Extracting this block does not refit the tracker at rank $q$. The prediction and service comparisons retain their different training, comparison-rank and returned-rank conventions in \cref{sec:prediction,sec:services}.

At equal rank, extraction may still rotate coordinates without changing the span, its objective or projector. The coordinate-standardized penalized probes in \cref{sec:prediction} can depend on that rotation. Lower-rank and cached query outputs are instead evaluated under the rank and freshness conventions in \cref{sec:services}.

For a fixed positive-semidefinite $M$, any rank-$q$ projector $P$, and a leading rank-$q$ projector $P_*^{(q)}$, assume $1\le q<D_x$, $J_q^*=\sum_{j\le q}\lambda_j(M)>0$ and $\delta_q(M)=\lambda_q(M)-\lambda_{q+1}(M)>0$. Then
\begin{equation}\label{eq:empirical_gap_projector}
\begin{aligned}
J_q^*-\tr(MP)&\ge\frac{\delta_q(M)}2\|P-P_*^{(q)}\|_F^2,\\
\|P-P_*^{(q)}\|_F&\le\sqrt{\frac{2J_q^*(1-\rho_q)}{\delta_q(M)}}.
\end{aligned}
\end{equation}
Here $\rho_q=\tr(MP)/J_q^*$. This empirical gap is distinct from a population gap or numerical-rank threshold; at a tied boundary the division bound is unavailable. A small relative gap permits high objective capture with substantial projector discrepancy. In the nonzero binary rank-one case, $\rho_1=\cos^2\theta$ and $\|P-P_*^{(1)}\|_F^2=2(1-\rho_1)$, linking the mean squared distance, not the mean distance, to mean objective capture. Proofs are in \suppref{supp:proofs}.

\subsection{Fixed-map theory and computational complexity}\label{sec:theory}

The following statement collects standard estimator, concentration and perturbation arguments with their separate conditions. It concerns any exact leading-$k$ spectral projector $\widehat P_t$ of $M_t=\bar C_t\bar C_t^\top$, and the leading-$k$ population projector $P_*$ of $M_*$.

Finite second moments suffice for consistency. Almost-sure bounds on whole feature vectors supply the stronger concentration statement. Identification of a particular leading population subspace additionally needs a positive separator after the selected rank. These conditions serve different purposes: neither bounded RFF outputs nor the centered-label rank cap alone supplies the population gap. Likewise, the pathwise recursion above remains correct for dependent or evolving-map features even when the probabilistic conclusions do not apply.

\begin{proposition}[Conditional estimator and exact-subspace guarantees]\label{prop:fixed_map_foundation}\label{prop:estimator_consistency}\label{prop:bounded_concentration}\label{prop:subspace_perturbation}
Condition on sampled map parameters and fixed preprocessing constants. If the feature pairs are i.i.d. with finite second moments, then
\begin{equation}\label{eq:fixed_map_consistency}
\E\bar C_t=(t-1)C_*/t,\qquad \bar C_t\longrightarrow C_*\quad\text{almost surely}.
\end{equation}
If also $\|\phi_t\|_2\le B_x$ and $\|\psi_t\|_2\le B_y$ almost surely, then for each fixed integer $t\ge1$ and $\delta\in(0,1)$, with probability at least $1-\delta$,
\begin{equation}\label{eq:fixed_map_concentration}
\opnorm{\bar C_t-C_*}\le b_t(\delta)
=6\sqrt2 B_xB_y\sqrt{\log(6/\delta)/t}.
\end{equation}
Separately, if $1\le k<D_x$ and the population boundary gap $\Delta_k=\lambda_k(M_*)-\lambda_{k+1}(M_*)$ is positive, then deterministically
\begin{equation}\label{eq:fixed_map_perturbation}
\begin{aligned}
\|\widehat P_t-P_*\|_F&\le\frac{2\sqrt{2k}}{\Delta_k}\opnorm{M_t-M_*},\\
\opnorm{M_t-M_*}&\le\opnorm{\bar C_t-C_*}\{2\opnorm{C_*}+\opnorm{\bar C_t-C_*}\}.
\end{aligned}
\end{equation}
\end{proposition}

The full proof in \suppref{supp:proofs} retains the centering term, fixed-time event and population-gap conditions \citep{pinelis1994,DavisKahan1970,Yu2015DKvariant}. Combining the bounds gives a leading $O(t^{-1/2})$ high-probability order for the exact projector at fixed dimensions, confidence and positive gap. RFF maps satisfy $\|\phi\|_2^2\le2$, and similarly for regression targets; one-hot targets have norm one. The concentration bound has no explicit feature-width factor, although embedding bounds, selected rank and eigengap can depend on dimension. Bounded features neither make evolving preprocessing i.i.d. nor control full-kernel approximation. The bounded fixed-feature study in \cref{sec:population} provides a finite-grid diagnostic, not an estimated convergence rate.

The three objects remain distinct throughout the empirical sections: the population projector belongs to a known fixed-feature distribution, the exact empirical projector belongs to its accumulated sample moment, and the practical projector belongs to the implemented recursion. Distances between these objects compare different subspace pairs; they do not form an additive attribution of observed error. The fixed-time probability statement likewise does not certify every requested basis along an adaptive or arbitrarily long stream.

With dimensions fixed, the core named arrays occupy $O(D_xD_y+3D_xk+D_x+D_y)$ entries, plus dense maps, scalers and implementation buffers specified in \suppref{supp:proofs}. Per-observation work has order
\begin{equation}\label{eq:update_complexity}
O(D_xd_x+I_{\rm reg}D_yd_y+D_xD_yk+D_xk^2+k^3),
\end{equation}
where $I_{\rm reg}$ indicates regression. Lower-order scaling, one-hot, mean and adaptive-array operations are included in this bound for positive dimensions and $k\ge1$. Associative multiplication avoids a $D_x\times D_x$ matrix. These arithmetic and named-array bounds are not process-memory or measured runtime laws; \cref{sec:services} charges the work needed to deliver a requested basis.

\section{Predictive performance and same-target diagnostics}\label{sec:prediction}

\subsection{Experimental setup}

The predictive comparison asks whether a more accurate solution of the empirical target explains the observed differences between representation pipelines. Four classification tasks use held-out macro-F1, and two regression tasks use held-out $R^2$. All reduced methods are trained at $q_{\rm cap}=C-1$ for classification and at $k=4$ for regression (\cref{tab:mt1}). The cap is an algebraic upper bound, not an estimated population rank or an optimal budget for every method. Ten final seeds, $100,\ldots,109$, share data, splits, and stream permutations across methods. Synthetic seeds change the generated data; real-data seeds preserve held-out membership. Their results are not pooled into a common task ranking.

\begin{table}[!tbp]
\centering
\caption{Predictive tasks and representation dimensions. Counts are final training/test sizes. The last column is the classification training cap $q_{\rm cap}=C-1$ or regression rank four. Original budgets for the unreduced-control comparison are stated separately in \suppref{tab:st4}.}
\label{tab:mt1}
\begingroup\setlength{\tabcolsep}{3pt}\renewcommand{\arraystretch}{1.08}
\begin{tabular*}{\linewidth}{@{\extracolsep{\fill}}lrrrrrr@{}}
\toprule
Task & $n_{\rm train}$ & $n_{\rm test}$ & $d_x$ & $D_x$ & $D_y$ & $k$ \\
\midrule
XOR-Gauss & 4,000 & 1,000 & 20 & 64 & 2 & 1 \\
Spirals-3 & 4,000 & 1,000 & 20 & 64 & 3 & 2 \\
Friedman-1 & 4,000 & 1,000 & 10 & 128 & 128 & 4 \\
HAR & 7,352 & 2,947 & 561 & 1,024 & 6 & 5 \\
Sensorless & 46,807 & 11,702 & 48 & 256 & 11 & 10 \\
Kin8nm & 6,553 & 1,639 & 8 & 128 & 128 & 4 \\
\bottomrule
\end{tabular*}
\endgroup

\end{table}

HAR preserves its official subject-disjoint split; Sensorless uses a fixed stratified split, and Kin8nm preserves its existing held-out partition \citep{UCIHAR2013,UCISensorless2013,OpenMLKin8nm}. These are benchmarks rather than application case studies. Synthetic generators, development partitions, bandwidths, optimizer settings, and uncertainty conventions are specified in \suppref{supp:prediction}.

After fitting the representation, prediction uses
\begin{equation}\label{eq:terminal_probe_pipeline}
z_U(x)=U^\top\phi(S_T^x(x)),\qquad
\widehat y_U(x)=\widehat h_U\!\left(S_U(z_U(x))\right).
\end{equation}
Here $S_U$ is fitted only to replayed training embeddings, and $\widehat h_U$ is logistic regression with inverse regularization one or Ridge with penalty one. Every representation receives its own standardizer and probe. Thus scores depend on coordinates, coordinatewise scaling, and penalized prediction, and are generally not rotation invariant. Terminal replay is retrospective: its fixed $S_T^x$ differs from the preceding scaler states that formed the causal moment. Test observations update neither representation nor probe.

OKSPCA and RFF-CIPLS share input RFF coordinates. For multiclass tasks, the CIPLS comparators concatenate one-versus-rest reducers and Gaussian-compress their outputs to the requested dimension; the displayed OVR--RP names identify these adapters. Regression instead compares OKSPCA's standardized RFF target with a scalar-target PLS recursion. CIPLS uses raw inputs, while CCIPCA and RFF-CCIPCA are unsupervised trackers. OKSPCA received development selection, whereas comparator settings were inherited without separate grids; the primary rank-cap fits were not retuned. These are complete-pipeline comparisons, not isolated interventions on optimization, target mapping, or supervised rank.

Exact replacement holds the accumulated moment and its feature history fixed; it leaves possible mismatch with the final replay coordinates unresolved. It is a numerical reference, not a population oracle.

\subsection{Predictive performance and exact-target comparisons}

\begin{figure}[!tbp]
\centering
\includegraphics[width=\textwidth]{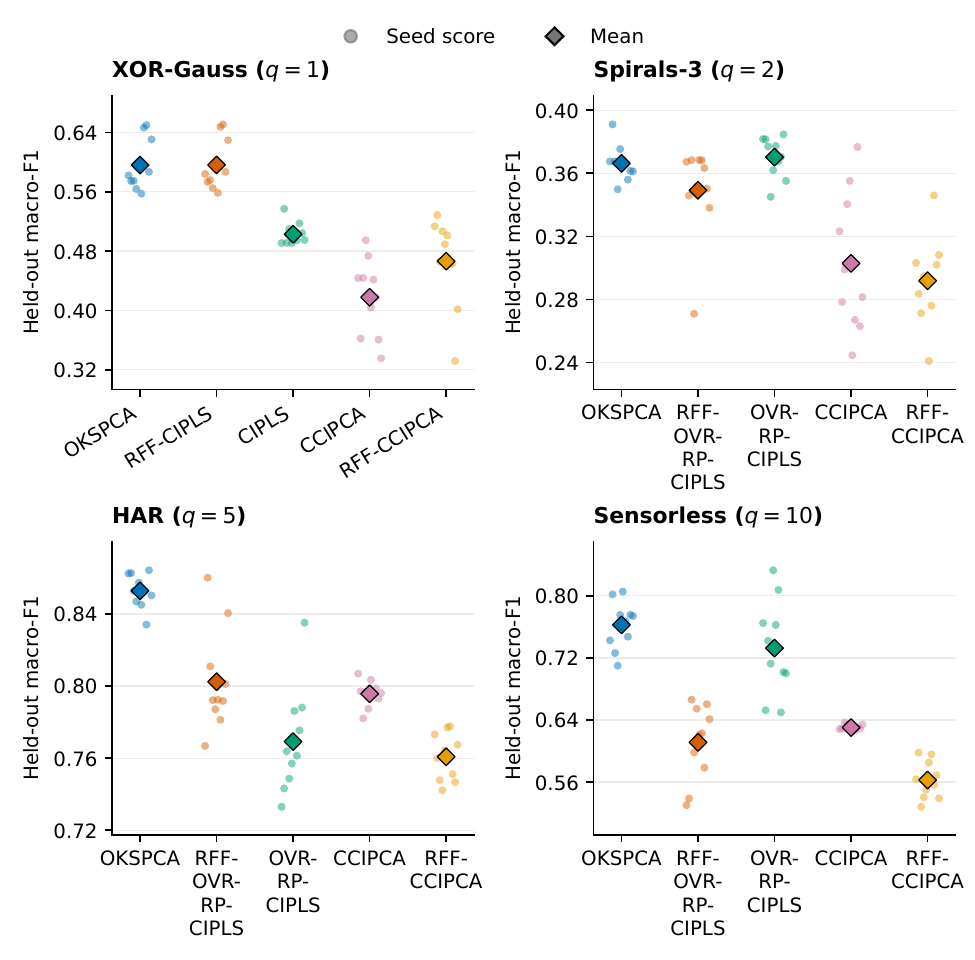}
\caption{Classification at the centered-label rank cap. Small points show all ten final-seed macro-F1 scores; diamonds show means. All five pipelines were trained from the beginning at $q_{\rm cap}=C-1$ without retuning. Multiclass OVR--RP labels distinguish the adapters from binary scalar-target CIPLS.}
\label{fig:mf1}
\end{figure}

\Cref{fig:mf1} retains all four classification tasks and five reduced pipelines. OKSPCA has the highest mean macro-F1 on HAR and Sensorless. Its mean differences from RFF--OVR--RP--CIPLS are $+0.0504$ and $+0.1513$, respectively. On Spirals-3 its corresponding difference is $+0.0170$, but OVR--RP--CIPLS has the higher mean. On binary XOR-Gauss the difference from RFF-CIPLS is $-0.0002$, with a fixed-seed interval spanning zero. The paired intervals in \suppref{tab:st3} describe the declared seeds and do not establish a general multiclass advantage.

On both regression tasks, OKSPCA scores below RFF-CIPLS, although above the two unsupervised trackers; linear CIPLS has the highest mean on Friedman-1 (\cref{tab:mt2}). Solving the same terminal empirical target exactly leaves these deficits largely unchanged. The paired exact-minus-full mean $R^2$ changes are $+0.000141$ for Friedman-1 and $+0.000027$ for Kin8nm. These are descriptive contrasts, not equivalence tests. They weaken the explanation that terminal tracking error alone causes the observed disadvantage, without identifying a unique alternative among feature mapping, target construction, output budget, and the evaluator.

\begin{table}[!tbp]
\centering
\caption{Regression at $k=4$. The reduced-method benchmark reports ten-seed $R^2$ means and sample standard deviations. The same-target comparison distinguishes the full maintained representation, its extracted block, and the exact empirical target; their absolute means accompany the descriptive paired exact-minus-full changes. Extracted-minus-exact intervals retain their separate identity in \suppref{tab:st3}.}
\label{tab:mt2}
\begingroup\setlength{\tabcolsep}{3pt}\renewcommand{\arraystretch}{1.08}
\begin{tabular*}{\linewidth}{@{\extracolsep{\fill}}lrr@{}}
\toprule
Reduced pipeline & Friedman-1 & Kin8nm \\
\midrule
OKSPCA & $0.6630 \pm 0.0417$ & $0.4130 \pm 0.0247$ \\
RFF-CIPLS & $0.7194 \pm 0.0242$ & $0.4546 \pm 0.0236$ \\
CIPLS & $0.7224 \pm 0.0161$ & $0.4043 \pm 0.0002$ \\
CCIPCA & $0.3012 \pm 0.1298$ & $0.1828 \pm 0.0979$ \\
RFF-CCIPCA & $0.2893 \pm 0.1150$ & $0.1747 \pm 0.0957$ \\
\bottomrule
\end{tabular*}
\par\medskip
\begin{tabular*}{\linewidth}{@{\extracolsep{\fill}}lrrrr@{}}
\toprule
Task & Full & Extracted & Exact & Exact $-$ full \\
\midrule
Friedman-1 & 0.662962 & 0.662963 & 0.663103 & +0.000141 \\
Kin8nm & 0.413004 & 0.413005 & 0.413032 & +0.000027 \\
\bottomrule
\end{tabular*}
\endgroup

\end{table}

\begin{samepage}
These regression endpoints originate from the larger-span terminal study, not from classification-rank reconstructions. Full, extracted, and exact representations have $k=q_T=4$, but extraction rotates coordinates before separate standardization and probe fitting. Their small observed differences do not make the evaluator rotation invariant. \suppref{supp:diagnostics} defines this comparison and reports its geometry. Unreduced Raw-X and RFF-X controls also retain an important limitation: RFF-X has higher original-budget mean scores than OKSPCA on five tasks, except Spirals-3 (\suppref{tab:st4}). Their different dimensions preclude reduced-method rankings.
\par\end{samepage}

Exact-minus-full measures replacement of the benchmark representation; extracted-minus-exact compares its within-span maximizing block with the unrestricted empirical solution. Only the latter has the intervals in \suppref{tab:st3}. Neither is a rotation-free subspace score or an equivalence test.

\subsection{Primary-rank classification diagnostics}

Forty primary-rank OKSPCA fits were reconstructed with the original inputs and settings, reproducing their terminal scores over all ten final seeds. Five retained checkpoints per trajectory supply 200 same-moment spectral comparisons. At termination, task-mean objective ratios range from $0.998824$ to $0.999951$; original and exact representations have similar mean scores under the stated probe (\cref{tab:mt3}). This compares directly trained ranks, not extraction from an earlier larger-rank classification fit.

\begin{table}[!htbp]
\centering
\caption{Direct primary-rank classification fits and their own exact terminal targets. Entries are means over ten seeds. Original/exact columns are macro-F1; $\rho$, maximum principal angle, and unnormalized Frobenius projector distance compare the retained state with the same empirical moment. These are terminal summaries, not uniform pathwise guarantees.}
\label{tab:mt3}
\begingroup\setlength{\tabcolsep}{3pt}\renewcommand{\arraystretch}{1.08}
\begin{tabular*}{\linewidth}{@{\extracolsep{\fill}}lrrrrrr@{}}
\toprule
Task & $k$ & Original $F_1$ & Exact $F_1$ & $\rho$ & $\theta_{\max}$ (deg) & $d_P$ \\
\midrule
XOR-Gauss & 1 & 0.5960 & 0.5962 & 0.999745 & 0.882 & 0.0218 \\
Spirals-3 & 2 & 0.3663 & 0.3648 & 0.998824 & 2.373 & 0.0683 \\
HAR & 5 & 0.8528 & 0.8527 & 0.999539 & 1.462 & 0.0448 \\
Sensorless & 10 & 0.7625 & 0.7625 & 0.999951 & 0.459 & 0.0151 \\
\bottomrule
\end{tabular*}
\endgroup

\end{table}

Terminal agreement is not uniform along the observed path. Sensorless seed 104 reaches a maximum angle of $78.099^\circ$ at observation 23,404, with $\rho=0.970687$. Its five-checkpoint record is retained in \suppref{tab:st5}; the finite checkpoints establish neither cause nor duration of the deviation. The reconstructed states are distinct from recovered historical states. Together, these diagnostics separate target capture, geometric agreement, and predictive usefulness while retaining the limitations of the original pipelines.

\FloatBarrier

\section{Empirical and population subspace accuracy}\label{sec:population}

The preceding comparisons have an empirical reference but no known population projector. A bounded fixed-feature experiment distinguishes these objects under the conditions of \cref{prop:fixed_map_foundation}. Its known cross-covariance $C_*$ has four positive squared singular values $(0.49,0.36,0.25,0.15)$; the selected subspace has rank three in dimensions $(D_x,D_y)=(32,8)$. Twenty seeds use nested sample sizes $256,512,\ldots,32{,}768$, with fixed population orientations. \suppref{supp:population} gives the generator and complete conventions. This setting does not reproduce the evolving preprocessing or random-feature approximation of the predictive benchmarks.

Let $P_*$, $P_{\rm exact,n}$, and $P_{\rm online,n}$ denote the population, exact empirical, and practical projectors. The three distances are
\begin{align}\label{eq:population_distances}
d_{\rm stat}&=\|P_{\rm exact,n}-P_*\|_F,\qquad
d_{\rm opt}=\|P_{\rm online,n}-P_{\rm exact,n}\|_F,\nonumber\\
d_{\rm total}&=\|P_{\rm online,n}-P_*\|_F.
\end{align}
They compare different subspace pairs and are not an additive error decomposition; $d_{\rm total}-d_{\rm stat}$ can have either sign. Objective ratios likewise depend on whether the reference operator is empirical or population.

\begin{figure}[!htbp]
\centering
\includegraphics[width=\textwidth]{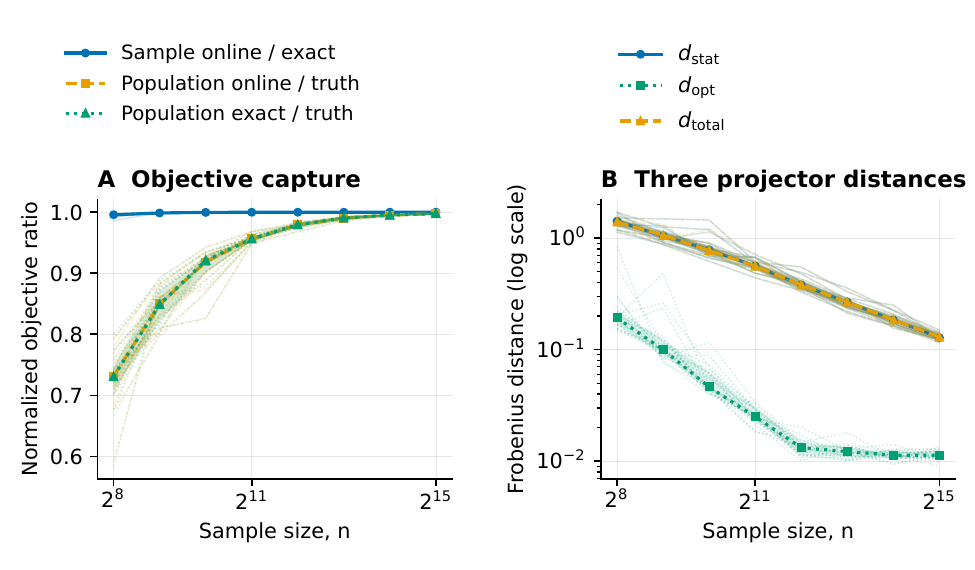}
\caption{Objective capture and projector distances over eight nested sample sizes in the bounded generator. The objective panel distinguishes sample-online, population-online, and population-exact ratios; the distance panel retains $d_{\rm stat}$, $d_{\rm opt}$, and $d_{\rm total}$. Thin traces show all 20 seeds, emphasized lines and bands their medians and interquartile ranges. The distances are not additive.}
\label{fig:mf2}
\end{figure}

At $n=256$, median sample-online objective capture is $0.995834$, while median $d_{\rm stat}$ is $1.416004$ and the population objective ratios are about $0.731$. Near-optimal empirical objective values therefore coexist with appreciable population error. Across the sample-size grid, median $d_{\rm stat}$ decreases to $0.1281$ and median $d_{\rm total}$ decreases from $1.395$ to $0.1292$. At the final checkpoint, median $d_{\rm opt}=0.01128$.

The sample-online ratio is normalized by the empirical optimum; both population ratios use the known population optimum. Thus near-one sample capture and population error concern different references, not fractions of predictive information retained. The finite-grid contraction is compatible with the fixed-map analysis, but estimates no convergence rate and establishes no convergence of the practical update. The bands condition on fixed population orientations while varying streams and tracker initialization. This experiment cannot attribute the predictive deficits in \cref{sec:prediction} to population estimation.

\FloatBarrier

\section{Computational trade-offs in basis access}\label{sec:services}

\subsection{Service definitions and experimental setup}

The service experiment asks what computation is required to deliver a basis for the current finite-feature moment. It has no predictive endpoint. Five synthetic workloads combine raw and feature dimensions, rank caps, and stream lengths as specified in \cref{tab:mt4}; R128 and R256 name target-feature widths, not raw response dimensions or predictive datasets. Five final seeds and five request schedules give 825 applicable method--regime--seed--schedule cells. All methods within a regime and seed process the same raw stream and causal frontend; dimensions vary jointly across regimes, so between-regime contrasts do not isolate target width.

\begin{table}[!htbp]
\centering
\caption{Numerical-service workloads. Raw responses are scalar labels or regression values. All applicable services use final seeds 6200--6204 and request periods 1, 10, 100, 1000, or Bernoulli probability 0.01, with termination included. Periodic caches retain separate update periods and receive no forced terminal refresh.}
\label{tab:mt4}
\begingroup\setlength{\tabcolsep}{3pt}\renewcommand{\arraystretch}{1.08}
\begin{tabular*}{\linewidth}{@{\extracolsep{\fill}}lrrrrrr@{}}
\toprule
Regime & $d_x$ & $d_y$ & $D_x$ & $D_y$ & $k$ & $n$ \\
\midrule
C2 & 20 & 1 & 64 & 2 & 1 & 4096 \\
C4 & 64 & 1 & 256 & 4 & 3 & 4096 \\
C11 & 48 & 1 & 256 & 11 & 10 & 4096 \\
R128 & 10 & 1 & 128 & 128 & 4 & 4096 \\
R256 & 64 & 1 & 256 & 256 & 8 & 2048 \\
\bottomrule
\end{tabular*}
\endgroup

\end{table}
\FloatBarrier

The offline full-moment reference defines
\begin{equation}\label{eq:service_rank}
r_{\rm ref}(t)=\#\{j:\sigma_j(\bar C_t)^2>10^{-10}\sigma_1(\bar C_t)^2\},
\qquad q_{\rm ref}(t)=\min\{k,r_{\rm ref}(t)\}.
\end{equation}
A zero leading singular value gives $r_{\rm ref}=q_{\rm ref}=0$. An approximate service applies the same relative squared-spectrum rule to $U_t^\top\bar C_t$, obtains $r_{\rm local}$, and returns $q_{\rm out}=\min(k,r_{\rm local})$ columns. It receives no offline full-spectrum rank; zero local rank does not establish a zero full moment. A request factors this small matrix and constructs its leading within-span block, including signed QR, copying, and checks. This paid query work leaves the maintained basis, moments, scalers, and optimizer unchanged; same-timestamp factors may be reused.

Six services apply throughout: Adam-style maintenance, normalized projected gradient (PG), full thin SVD after each observation, full thin SVD on request, and exact caches updated every ten or one hundred observations. Binary direct extraction additionally applies to C2; contrast QR followed by a small SVD applies to C4 and C11. Generic exact services compute the full thin factorization before truncation. Structured services exploit centered-class identities, with floating-point structural residuals measured separately. A cache returns its last available basis and timestamp $\tau$, of age $t-\tau$; before initialization it returns no basis.

\begin{samepage}
For binary labels, $\bar C_t=[c_t,-c_t]$ gives $\bar C_t\bar C_t^\top=2c_tc_t^\top$ and exact direction $c_t/\|c_t\|_2$ when $c_t\ne0$. More generally, an orthonormal Helmert contrast $H_C$ spans $\mathbf1_C^\perp$. Centered labels imply $\bar C_tH_CH_C^\top\bar C_t^\top=\bar C_t\bar C_t^\top$. Factoring the smaller contrast matrix therefore preserves the target. A plain thin QR alone identifies the maximizing span only with sufficient input dimension, full contrast rank, and the full requested contrast span. The evaluated contrast service instead takes an SVD of its triangular factor to detect rank and select the required block (\suppref{supp:services}).
\par\end{samepage}

PG shares Adam's seeded initial basis but uses
\begin{equation}\label{eq:pg_service_update}
\begin{aligned}
H_t&=2C_t(C_t^\top U_{t-1}),\qquad
H_{T,t}=H_t-U_{t-1}\sym(U_{t-1}^\top H_t),\\
U_t&=\qf_+\!\left(U_{t-1}+\frac{\gamma H_{T,t}}{2\|\bar C_t\|_F^2+10^{-12}}\right).
\end{aligned}
\end{equation}
The gradient uses Bessel-normalized $C_t$, whereas the denominator uses $\bar C_t$. PG has no Adam arrays, clipping, or line search. An exactly zero tangent gives zero displacement followed by signed QR. Its four-candidate, three-seed development selection differs from Adam's fixed settings (\suppref{supp:services}).

Quality is defined only for valid outputs with $q_{\rm out}=q_{\rm ref}>0$, against the current moment. Zero-target and unequal-rank geometry remain undefined. A target boundary is flagged tied when its squared-spectrum gap is at most $10^{-12}\max\{1,\sigma_1^2\}$. This numerical flag includes near ties; an exactly tied boundary has an objective value but does not identify a unique projector. Shortfall counts positive-reference requests without valid same-rank output. Staleness overlaps these categories. These conventions differ from the direct-rank reconstruction conventions in \suppref{supp:diagnostics}.

Freshness certifies when a block was constructed, not its geometric accuracy. An old cache can satisfy the current-rank requirement at one request and fail it at another; denominators and stale counts therefore accompany conditional geometry.

\subsection{Request schedules and service time}

The primary interval A0 includes causal scaling, feature mapping, moments, maintenance, and requested output construction after initialization. Initialization is reported separately and explicitly added for an inclusive total. Stream preparation, offline references, untimed quality checks, hashing, and serialization are excluded. Separate pass B measures inclusive batch means and synchronous query durations; pass C checks quality. They neither decompose A0 additively nor provide extra input seeds, and the synchronous experiment has no external-arrival queue.

\Cref{tab:mt5} shows every regime--schedule combination. Entries are medians of five seed-paired Adam/exact-request A0 wall-time ratios, not ratios of medians. Exact-request is faster in all 75 classification seed pairs. Adam is faster for every-observation R128 requests and the two densest periodic R256 schedules. R128 period ten favors Adam in only three of five pairs; sparser regression requests favor exact-request at the group median. This grid does not establish a universal crossover frequency.

\begin{table}[!htbp]
\centering
\caption{All 25 request-schedule comparisons. Values below one favor Adam; each entry is the median of five paired A0 wall-time ratios against exact-request. R128 period ten favors Adam in only 3/5 pairs. The sparse R128 anomaly remains included; ranges, faster counts, and initialization-inclusive comparisons appear in \suppref{tab:st8}.}
\label{tab:mt5}
\begingroup\setlength{\tabcolsep}{3pt}\renewcommand{\arraystretch}{1.08}
\begin{tabular*}{\linewidth}{@{\extracolsep{\fill}}lrrrrr@{}}
\toprule
Regime & Every 1 & Every 10 & Every 100 & Every 1000 & Bernoulli \\
\midrule
C2 & 1.726 & 2.210 & 2.242 & 2.044 & 2.199 \\
C4 & 1.806 & 2.259 & 2.488 & 2.452 & 2.541 \\
C11 & 1.513 & 2.628 & 2.830 & 2.886 & 2.899 \\
R128 & 0.277 & 0.923 & 1.841 & 2.097 & 1.719 \\
R256 & 0.085 & 0.567 & 1.625 & 1.622 & 1.435 \\
\bottomrule
\end{tabular*}
\endgroup

\end{table}
\FloatBarrier

\begin{samepage}
For R128 period 1000, seed 6204 has exact-request A0 wall time 3.225 seconds and process-CPU time 0.625 seconds. Its cause is unresolved; it remains in the ranges and faster counts. Prescribed identical-workload repeats and accumulation-only controls are retained in \suppref{tab:st9}; they do not establish grid-wide timing stability.
\par\end{samepage}

\subsection{Subspace quality, coverage, and staleness}

The joint every-observation comparison covers 33 applicable method--regime combinations, a different grid from the 25 schedule contrasts. \Cref{tab:mt6} shows both regression workloads; \suppref{tab:st10} and \suppref{tab:st11} contain the complete classification and regression results. Conditional quality averages valid matched-rank requests within each seed, then weights the five seed means equally. Coverage and stale counts pool requests; they are not independent replications. Absolute reference times are directly summarized measured times, not products of ratio medians.

\begin{table}[!htbp]
\centering
\caption{Every-observation service in R128 and R256. Time is the median paired A0 ratio against exact-request; headings give its absolute median. Conditional geometry averages within seed before equal seed weighting. Matched/positive-reference and shortfall count coverage; stale overlaps other counts. Both cache periods and PG remain visible beside Adam.}
\label{tab:mt6}
\begingroup\setlength{\tabcolsep}{3pt}\renewcommand{\arraystretch}{1.08}
\begin{tabular*}{\linewidth}{@{\extracolsep{\fill}}lrrrrrrr@{}}
\toprule
Service & \shortstack{Paired A0\\ratio} & Mean $\rho$ & Mean $d_P$ & \shortstack{Mean angle\\(deg)} & \shortstack{Matched /\\positive} & Shortfall & Stale \\
\midrule
\multicolumn{8}{l}{\textit{R128: Exact-request A0 wall median = 5,642 ms}} \\
\shortstack[l]{Exact-\\request} & 1.000 & 1.0000 & 0 & 1.8e-06 & 20475/20475 & 0 & 0 \\
Adam & 0.277 & 0.9905 & 0.658 & 28.83 & 20475/20475 & 0 & 0 \\
PG & 0.280 & 0.9971 & 0.447 & 17.11 & 20475/20475 & 0 & 0 \\
Cache-10 & 0.193 & 0.9979 & 0.0951 & 3.48 & 20435/20475 & 40 & 18390 \\
Cache-100 & 0.107 & 0.9930 & 0.299 & 9.92 & 19985/20475 & 490 & 19785 \\
\addlinespace[4pt]
\multicolumn{8}{l}{\textit{R256: Exact-request A0 wall median = 18,840 ms}} \\
\shortstack[l]{Exact-\\request} & 1.000 & 1.0000 & 0 & 2.2e-06 & 10235/10235 & 0 & 0 \\
Adam & 0.085 & 0.9719 & 2.31 & 87.67 & 10210/10235 & 25 & 0 \\
PG & 0.113 & 0.9201 & 2.98 & 87.94 & 10226/10235 & 9 & 0 \\
Cache-10 & 0.134 & 0.9931 & 0.252 & 8.67 & 10187/10235 & 48 & 9175 \\
Cache-100 & 0.047 & 0.9612 & 0.815 & 23.28 & 9745/10235 & 490 & 9645 \\
\bottomrule
\end{tabular*}
\endgroup

\end{table}
\FloatBarrier

In R128, Cache-10 combines lower cost and higher conditional quality than Adam with 40 shortfalls and 18,390 stale outputs. In R256, Adam's time ratio is 0.085 but mean $d_P=2.31$ and mean maximum angle is $87.67^\circ$. PG has higher equally seed-weighted full-stream mean objective capture than Adam in C2, C4, C11, and R128, but lower capture in R256; R128 endpoint ordering differs from stream-average ordering. The prescribed shortfall-first rule selected $\gamma=0.1$ for R256 despite another candidate's smaller objective-loss average.

Contrast extraction is slower than generic exact-request in C4 and slightly faster in C11, while retaining near-unit objective capture and very small geometric residuals. Cache-10 and Cache-100 cost less than Adam in all three classification regimes, but include stale returns and early shortfalls. These orderings concern the measured implementations and full service boundaries, not complexity alone.

\begin{samepage}
Across schedules, 689,028 requests include 688,828 positive references and 685,784 valid matched-rank outputs. The 3,044 shortfalls comprise 2,915 missing valid outputs and 129 insufficient-rank returns; 181,473 stale responses overlap these categories. Undefined geometry is not replaced by zero. Thus low service time may describe a missing, stale, or rank-limited response.
\par\end{samepage}

\begin{figure}[!tbp]
\centering
\includegraphics[width=\textwidth]{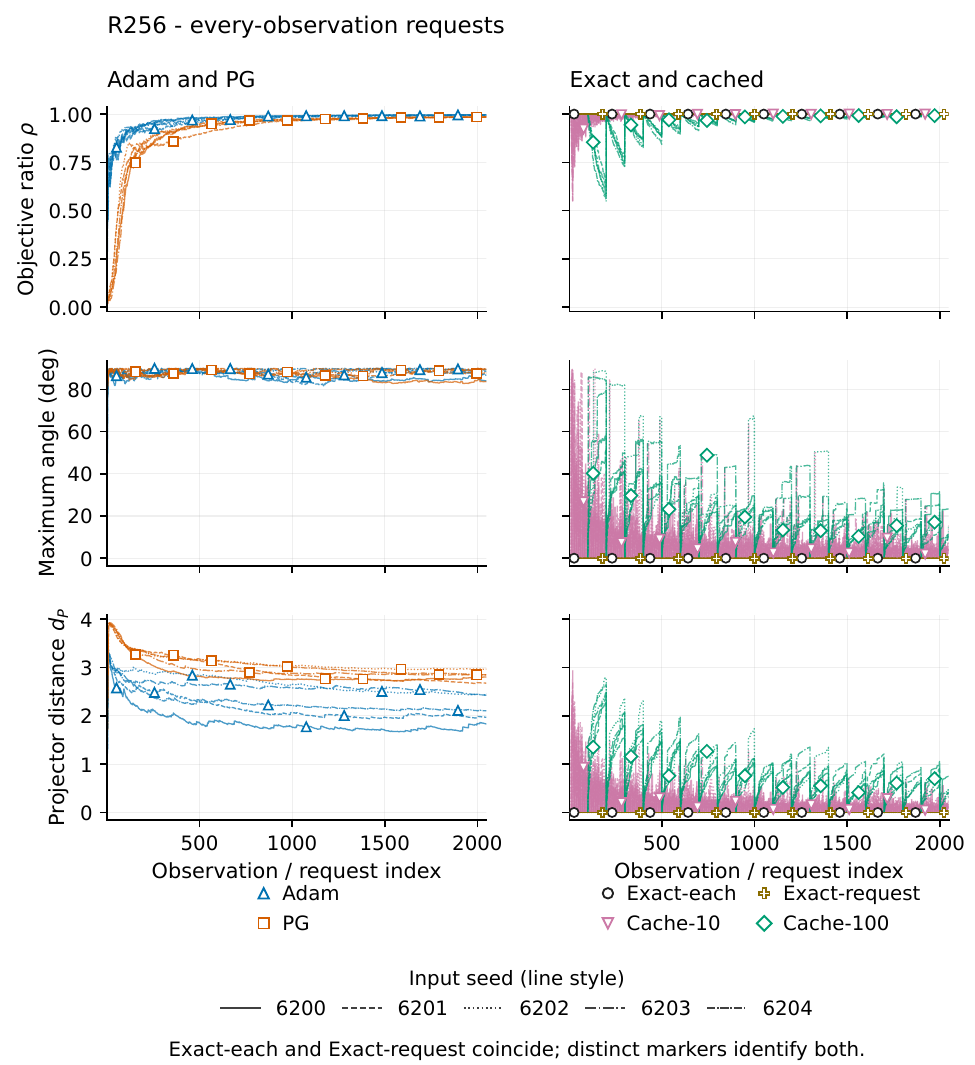}
\caption{Complete R256 every-observation trajectories for all five seeds. Columns separate Adam/PG from exact/cached services on identical axes; rows show objective ratio, maximum principal angle, and projector distance. Undefined matched-rank metrics remain gaps. Line styles identify seeds (6202 is dotted) and colours/markers identify services. R128's distinct full-stream behavior is shown in \suppref{fig:sf1}.}
\label{fig:mf3}
\end{figure}
\FloatBarrier

\begin{samepage}
R256's geometric discrepancy persists beyond its early low-objective period (\cref{fig:mf3}). All 5,120 recorded Adam angles at observations 1025--2048 exceed $80^\circ$, with valid matched-rank, fresh outputs and no flagged target tie. This window and threshold are post hoc descriptions. The separate saved-state correspondence covers five terminal responses, not all scalar requests (\suppref{tab:st12}). Maximum angle concerns the least-aligned direction; projector distances also indicate aggregate error. Small relative boundary gaps make such geometry compatible with high objective capture through \cref{eq:empirical_gap_projector}, without identifying an optimizer failure or establishing superiority over independent leading-$k$ services.
\par\end{samepage}

\section{Discussion}\label{sec:discussion}
\subsection{Interpretation of the findings}
The same-target intervention separates numerical accuracy from the predictive deficit: improving the terminal solution leaves the regression disadvantage relative to RFF-CIPLS. It narrows an optimization-based explanation without selecting a unique cause among target encoding, feature mapping, coordinate scaling and the separately fitted probe.

The intermediate-state and population diagnostics identify what this intervention leaves open. Terminal summaries miss an observed intermediate deviation, while empirical agreement does not establish population recovery. Neither the finite sample-size grid nor the saved reconstruction checkpoints establish a limiting practical discrepancy or convergence rate.

\subsection{Scope and limitations}
The cost advantage of maintenance is conditional on the requested output. R256 illustrates a fresh, matched-rank service with substantial geometric error; R128's fast, conditionally accurate Cache-10 instead incurs stale returns and rank shortfalls. Neither low time nor nominal coverage can stand in for current-subspace quality. Relative gaps explain compatibility between energy and geometry, not the cause of an optimizer's behavior.

An important computational alternative remains untested. Avoiding repeated full-spectrum computation need not require the Adam update. Projected gradient already supplies a warm-started small-block method, but does not replace a comparison against a distinct leading-$k$ solver with a specified work subspace and budget or an incremental low-rank update. The present measurements do not establish either alternative's accuracy or speed. Instrumented batch means are not individual-update tail latency, and synchronous responses are not deployment queueing measurements.

The prediction studies retain further limitations. Multiclass CIPLS uses OVR concatenation and random projection; changing the rank cap changes both operations. Shared input features do not remove target-construction differences, coordinatewise evaluation or asymmetric tuning. The numerical-service comparisons do not resolve those predictive confounders, and PG's development-selected step differs from Adam's fixed setting. The fixed-map theory also excludes the coupled effects of evolving standardizers and the actual adaptive update. Existing stochastic and Riemannian analyses use different assumptions or constructions \citep{bonnabel2013sgd,Vary2024RandomStiefel,SakaiIiduka2025Riemannian}. The six predictive tasks and bounded numerical workloads are not application deployments or concept-drift studies \citep{Bienstock2022RobustStreamingPCA}.

The resulting diagnosis is conditional: the target, its practical solution and the requested service require separate evaluation. Their agreement cannot be presumed from any single predictive, objective or timing summary, nor do these comparisons select a universally preferable method.

\paragraph{Reproducibility.}
The arXiv source package contains the manuscript source and publication assets required to build this paper. A separate local package contains frozen display inputs and rendering scripts, but is not publicly archived at the time of this version. These materials do not constitute a complete experimental rerun package.

\begingroup
\small\singlespacing
\bibliographystyle{unsrtnat}
\bibliography{references}

\begin{thebibliography}{44}
\providecommand{\natexlab}[1]{#1}
\providecommand{\url}[1]{\texttt{#1}}
\expandafter\ifx\csname urlstyle\endcsname\relax
  \providecommand{\doi}[1]{doi: #1}\else
  \providecommand{\doi}{doi: \begingroup \urlstyle{rm}\Url}\fi

\bibitem[Bair et~al.(2006)Bair, Hastie, Paul, and Tibshirani]{Bair2006SPC}
Eric Bair, Trevor Hastie, Debashis Paul, and Robert Tibshirani.
\newblock Prediction by supervised principal components.
\newblock \emph{Journal of the American Statistical Association}, 101\penalty0
  (473):\penalty0 119--137, 2006.
\newblock \doi{10.1198/016214505000000628}.

\bibitem[Barshan et~al.(2011)Barshan, Ghodsi, Azimifar, and
  Zolghadri~Jahromi]{Barshan2011SPCA}
Elnaz Barshan, Ali Ghodsi, Zohreh Azimifar, and Mansoor Zolghadri~Jahromi.
\newblock Supervised principal component analysis: Visualization,
  classification and regression on subspaces and submanifolds.
\newblock \emph{Pattern Recognition}, 44\penalty0 (7):\penalty0 1357--1371,
  2011.
\newblock \doi{10.1016/j.patcog.2010.12.015}.

\bibitem[Papazoglou and Yin(2025)]{PapazoglouYin2025CSPCA}
Theodosios Papazoglou and Guosheng Yin.
\newblock Covariance supervised principal component analysis.
\newblock arXiv:2506.19247, 2025.
\newblock Preprint.

\bibitem[Ghosh and Kirby(2022)]{GhoshKirby2022Centroid}
Tomojit Ghosh and Michael Kirby.
\newblock Supervised dimensionality reduction and visualization using
  centroid-encoder.
\newblock \emph{Journal of Machine Learning Research}, 23\penalty0
  (20):\penalty0 1--34, 2022.
\newblock URL \url{https://www.jmlr.org/papers/v23/20-188.html}.

\bibitem[Gretton et~al.(2005)Gretton, Bousquet, Smola, and
  Sch{\"o}lkopf]{gretton2005}
Arthur Gretton, Olivier Bousquet, Alex Smola, and Bernhard Sch{\"o}lkopf.
\newblock Measuring statistical dependence with {Hilbert--Schmidt} norms.
\newblock In \emph{Algorithmic Learning Theory}, volume 3734 of \emph{Lecture
  Notes in Computer Science}, pages 63--77. Springer, 2005.
\newblock \doi{10.1007/11564089_7}.

\bibitem[Karimi et~al.(2018)Karimi, Wong, and Ghodsi]{Karimi2018SRP}
Amir-Hossein Karimi, Alexander Wong, and Ali Ghodsi.
\newblock {SRP}: Efficient class-aware embedding learning for large-scale data
  via supervised random projections, 2018.
\newblock URL \url{https://arxiv.org/abs/1811.03166}.
\newblock Preprint.

\bibitem[Hotelling(1936)]{Hotelling1936CCA}
Harold Hotelling.
\newblock Relations between two sets of variates.
\newblock \emph{Biometrika}, 28\penalty0 (3-4):\penalty0 321--377, 1936.
\newblock \doi{10.1093/biomet/28.3-4.321}.

\bibitem[Frank and Friedman(1993)]{FrankFriedman1993Chemometrics}
Ildiko~E. Frank and Jerome~H. Friedman.
\newblock A statistical view of some chemometrics regression tools.
\newblock \emph{Technometrics}, 35\penalty0 (2):\penalty0 109--135, 1993.
\newblock \doi{10.1080/00401706.1993.10485033}.

\bibitem[Rosipal and Trejo(2001)]{Rosipal2001KPLS}
Roman Rosipal and Leonard~J Trejo.
\newblock Kernel partial least squares regression in reproducing kernel
  {Hilbert} space.
\newblock \emph{Journal of Machine Learning Research}, 2\penalty0
  (Dec):\penalty0 97--123, 2001.
\newblock URL \url{https://www.jmlr.org/papers/v2/rosipal01a.html}.

\bibitem[Rahimi and Recht(2007)]{Rahimi2007RFF}
Ali Rahimi and Benjamin Recht.
\newblock Random features for large-scale kernel machines.
\newblock In \emph{Advances in Neural Information Processing Systems},
  volume~20, pages 1177--1184, 2007.

\bibitem[Dayal and MacGregor(1997)]{DayalMacGregor1997REWPLS}
Bhupinder~S. Dayal and John~F. MacGregor.
\newblock Recursive exponentially weighted {PLS} and its applications to
  adaptive control and prediction.
\newblock \emph{Journal of Process Control}, 7\penalty0 (3):\penalty0 169--179,
  1997.
\newblock \doi{10.1016/S0959-1524(97)80001-7}.

\bibitem[Qin(1998)]{Qin1998RPLS}
S.~Joe Qin.
\newblock Recursive {PLS} algorithms for adaptive data modeling.
\newblock \emph{Computers \& Chemical Engineering}, 22\penalty0 (4-5):\penalty0
  503--514, 1998.
\newblock \doi{10.1016/S0098-1354(97)00262-7}.

\bibitem[Wang et~al.(2003)Wang, Kruger, and Lennox]{wang2003recursive}
Xun Wang, Uwe Kruger, and Barry Lennox.
\newblock Recursive partial least squares algorithms for monitoring complex
  industrial processes.
\newblock \emph{Control Engineering Practice}, 11\penalty0 (6):\penalty0
  613--632, 2003.
\newblock \doi{10.1016/S0967-0661(02)00096-5}.

\bibitem[Zeng and Li(2014)]{Zeng2014IPLS}
Xue-Qiang Zeng and Guo-Zheng Li.
\newblock Incremental partial least squares analysis of big streaming data.
\newblock \emph{Pattern Recognition}, 47\penalty0 (11):\penalty0 3726--3735,
  2014.
\newblock \doi{10.1016/j.patcog.2014.05.022}.

\bibitem[Arora et~al.(2012)Arora, Cotter, Livescu, and
  Srebro]{arora2012stochastic}
Raman Arora, Andrew Cotter, Karen Livescu, and Nathan Srebro.
\newblock Stochastic optimization for {PCA} and {PLS}.
\newblock In \emph{2012 50th Annual Allerton Conference on Communication,
  Control, and Computing (Allerton)}, pages 861--868. IEEE, 2012.
\newblock \doi{10.1109/Allerton.2012.6483308}.

\bibitem[Arora et~al.(2016)Arora, Mianjy, and Marinov]{Arora2016SAPLS}
Raman Arora, Poorya Mianjy, and Teodor Marinov.
\newblock Stochastic optimization for multiview representation learning using
  partial least squares.
\newblock In \emph{Proceedings of the 33rd International Conference on Machine
  Learning}, volume~48 of \emph{Proceedings of Machine Learning Research},
  pages 1786--1794. PMLR, 2016.
\newblock URL \url{https://proceedings.mlr.press/v48/aroraa16.html}.

\bibitem[Chen et~al.(2017)Chen, Yang, Li, and Zhao]{Chen2017OnlinePLS}
Zhehui Chen, Lin~F. Yang, Chris~Junchi Li, and Tuo Zhao.
\newblock Online partial least square optimization: Dropping convexity for
  better efficiency and scalability.
\newblock In \emph{Proceedings of the 34th International Conference on Machine
  Learning}, volume~70 of \emph{Proceedings of Machine Learning Research},
  pages 777--786. PMLR, 2017.
\newblock URL \url{https://proceedings.mlr.press/v70/chen17h.html}.

\bibitem[Kaiser et~al.(2010)Kaiser, Schenck, and
  M{\"o}ller]{Kaiser2010CoupledSVD}
Alexander Kaiser, Wolfram Schenck, and Ralf M{\"o}ller.
\newblock Coupled singular value decomposition of a cross-covariance matrix.
\newblock \emph{International Journal of Neural Systems}, 20\penalty0
  (4):\penalty0 293--318, 2010.
\newblock \doi{10.1142/S0129065710002437}.

\bibitem[Feng et~al.(2017)Feng, Kong, Xu, and Qin]{Feng2017PSS}
Xiaowei Feng, Xiangyu Kong, Donghui Xu, and Jianqiang Qin.
\newblock A fast and effective principal singular subspace tracking algorithm.
\newblock \emph{Neurocomputing}, 267:\penalty0 201--209, 2017.
\newblock \doi{10.1016/j.neucom.2017.06.006}.

\bibitem[Xie et~al.(2015)Xie, Song, Dai, Li, and Song]{Xie2015OSDR}
Yao Xie, Ruiyang Song, Hanjun Dai, Qingbin Li, and Le~Song.
\newblock Online supervised subspace tracking.
\newblock arXiv:1509.00137, 2015.
\newblock Preprint.

\bibitem[Cai et~al.(2020)Cai, Li, and Zhu]{Cai2020OnlineSIR}
Zhanrui Cai, Runze Li, and Liping Zhu.
\newblock Online sufficient dimension reduction through sliced inverse
  regression.
\newblock \emph{Journal of Machine Learning Research}, 21\penalty0
  (10):\penalty0 1--25, 2020.
\newblock URL \url{https://www.jmlr.org/papers/v21/18-567.html}.

\bibitem[Xu et~al.(2025)Xu, Zhao, and Cheng]{Xu2025OnlineKSIR}
Jianjun Xu, Yue Zhao, and Haoyang Cheng.
\newblock Online kernel sliced inverse regression.
\newblock \emph{Computational Statistics \& Data Analysis}, 203:\penalty0
  108071, 2025.
\newblock \doi{10.1016/j.csda.2024.108071}.

\bibitem[Brand(2006)]{brand2006fast}
Matthew Brand.
\newblock Fast low-rank modifications of the thin singular value decomposition.
\newblock \emph{Linear Algebra and its Applications}, 415\penalty0
  (1):\penalty0 20--30, 2006.
\newblock \doi{10.1016/j.laa.2005.07.021}.

\bibitem[Oja(1982)]{Oja1982PCA}
Erkki Oja.
\newblock Simplified neuron model as a principal component analyzer.
\newblock \emph{Journal of Mathematical Biology}, 15\penalty0 (3):\penalty0
  267--273, 1982.
\newblock \doi{10.1007/BF00275687}.

\bibitem[Chin and Suter(2007)]{Chin2007IKPCA}
Tat-Jun Chin and David Suter.
\newblock Incremental kernel principal component analysis.
\newblock \emph{IEEE Transactions on Image Processing}, 16\penalty0
  (6):\penalty0 1662--1674, 2007.
\newblock \doi{10.1109/TIP.2007.896668}.

\bibitem[Ullah et~al.(2018)Ullah, Mianjy, Marinov, and
  Arora]{Ullah2018StreamingKPCA}
Enayat Ullah, Poorya Mianjy, Teodor~V. Marinov, and Raman Arora.
\newblock Streaming kernel {PCA} with $\tilde{O}(\sqrt{n})$ random features.
\newblock In \emph{Advances in Neural Information Processing Systems},
  volume~31, pages 7311--7321. Curran Associates, Inc., 2018.
\newblock URL
  \url{https://proceedings.neurips.cc/paper_files/paper/2018/hash/7ae11af20803185120e83d3ce4fb4ed7-Abstract.html}.

\bibitem[Deng et~al.(2025)Deng, Long, Song, Wang, and
  Zhang]{Deng2025StreamingKPCA}
Yichuan Deng, Jiangxuan Long, Zhao Song, Zifan Wang, and Han Zhang.
\newblock Streaming kernel {PCA} algorithm with small space.
\newblock In \emph{Conference on Parsimony and Learning}, volume 280 of
  \emph{Proceedings of Machine Learning Research}, pages 1216--1254. PMLR,
  2025.
\newblock URL \url{https://proceedings.mlr.press/v280/deng25a.html}.

\bibitem[Jord{\~a}o et~al.(2021)Jord{\~a}o, Lie, de~Melo, and
  Schwartz]{Jordao2021CIPLS}
Artur Jord{\~a}o, Maiko Lie, Victor Hugo~Cunha de~Melo, and William~Robson
  Schwartz.
\newblock Covariance-free partial least squares: An incremental dimensionality
  reduction method.
\newblock In \emph{Proceedings of the IEEE/CVF Winter Conference on
  Applications of Computer Vision (WACV)}, pages 1420--1428, 2021.
\newblock \doi{10.1109/WACV48630.2021.00146}.

\bibitem[Ashtiani and Ghodsi(2015)]{Ashtiani2015KSPCABound}
Hassan Ashtiani and Ali Ghodsi.
\newblock A dimension-independent generalization bound for kernel supervised
  principal component analysis.
\newblock In \emph{Proceedings of the 1st International Workshop on Feature
  Extraction: Modern Questions and Challenges at NIPS 2015}, volume~44 of
  \emph{Proceedings of Machine Learning Research}, pages 19--29. PMLR, 2015.
\newblock URL \url{https://proceedings.mlr.press/v44/Ashtiani2015.html}.

\bibitem[Fan(1949)]{fan1949}
Ky~Fan.
\newblock On a theorem of {W}eyl concerning eigenvalues of linear
  transformations {I}.
\newblock \emph{Proceedings of the National Academy of Sciences of the United
  States of America}, 35\penalty0 (11):\penalty0 652--655, 1949.
\newblock \doi{10.1073/pnas.35.11.652}.

\bibitem[Overton and Womersley(1992)]{overton1992sum}
Michael~L. Overton and Robert~S. Womersley.
\newblock On the sum of the largest eigenvalues of a symmetric matrix.
\newblock \emph{SIAM Journal on Matrix Analysis and Applications}, 13\penalty0
  (1):\penalty0 41--45, 1992.
\newblock \doi{10.1137/0613006}.

\bibitem[Absil et~al.(2008)Absil, Mahony, and Sepulchre]{Absil2008Manifolds}
P.-A. Absil, Robert Mahony, and Rodolphe Sepulchre.
\newblock \emph{Optimization Algorithms on Matrix Manifolds}.
\newblock Princeton University Press, 2008.

\bibitem[Kingma and Ba(2015)]{kingma2014adam}
Diederik~P Kingma and Jimmy Ba.
\newblock {Adam}: A method for stochastic optimization.
\newblock In \emph{International Conference on Learning Representations}, 2015.
\newblock URL \url{https://openreview.net/forum?id=8gmWwjFyLj}.

\bibitem[B{\'e}cigneul and Ganea(2019)]{Becigneul2018riemannian}
Gary B{\'e}cigneul and Octavian-Eugen Ganea.
\newblock {Riemannian} adaptive optimization methods.
\newblock In \emph{International Conference on Learning Representations}, 2019.
\newblock URL \url{https://openreview.net/forum?id=r1eiqi09K7}.

\bibitem[Bonnabel(2013)]{bonnabel2013sgd}
Silv\`ere Bonnabel.
\newblock Stochastic gradient descent on {Riemannian} manifolds.
\newblock \emph{IEEE Transactions on Automatic Control}, 58\penalty0
  (9):\penalty0 2217--2229, 2013.
\newblock \doi{10.1109/TAC.2013.2254619}.

\bibitem[Vary et~al.(2024)Vary, Ablin, Gao, and Absil]{Vary2024RandomStiefel}
Simon Vary, Pierre Ablin, Bin Gao, and Pierre-Antoine Absil.
\newblock Optimization without retraction on the random generalized {Stiefel}
  manifold.
\newblock In \emph{Proceedings of the 41st International Conference on Machine
  Learning}, volume 235 of \emph{Proceedings of Machine Learning Research},
  pages 49226--49248, 2024.
\newblock URL \url{https://proceedings.mlr.press/v235/vary24a.html}.

\bibitem[Sakai and Iiduka(2025)]{SakaiIiduka2025Riemannian}
Hiroyuki Sakai and Hideaki Iiduka.
\newblock A general framework of {R}iemannian adaptive optimization methods
  with a convergence analysis.
\newblock \emph{Transactions on Machine Learning Research}, 2025.
\newblock ISSN 2835-8856.
\newblock URL \url{https://openreview.net/forum?id=knv4lQFVoE}.

\bibitem[Pinelis(1994)]{pinelis1994}
Iosif Pinelis.
\newblock Optimum bounds for the distributions of martingales in {Banach}
  spaces.
\newblock \emph{The Annals of Probability}, 22\penalty0 (4):\penalty0
  1679--1706, 1994.
\newblock \doi{10.1214/aop/1176988477}.

\bibitem[Davis and Kahan(1970)]{DavisKahan1970}
Chandler Davis and William~M. Kahan.
\newblock The rotation of eigenvectors by a perturbation. {III}.
\newblock \emph{SIAM Journal on Numerical Analysis}, 7\penalty0 (1):\penalty0
  1--46, 1970.
\newblock \doi{10.1137/0707001}.

\bibitem[Yu et~al.(2015)Yu, Wang, and Samworth]{Yu2015DKvariant}
Yi~Yu, Tengyao Wang, and Richard~J. Samworth.
\newblock A useful variant of the {D}avis--{K}ahan theorem for statisticians.
\newblock \emph{Biometrika}, 102\penalty0 (2):\penalty0 315--323, 2015.
\newblock \doi{10.1093/biomet/asv008}.

\bibitem[Reyes-Ortiz et~al.(2013)Reyes-Ortiz, Anguita, Ghio, Oneto, and
  Parra]{UCIHAR2013}
Jorge Reyes-Ortiz, Davide Anguita, Alessandro Ghio, Luca Oneto, and Xavier
  Parra.
\newblock Human activity recognition using smartphones, 2013.
\newblock URL \url{https://doi.org/10.24432/C54S4K}.

\bibitem[Bator(2013)]{UCISensorless2013}
Martyna Bator.
\newblock Dataset for sensorless drive diagnosis, 2013.
\newblock URL \url{https://doi.org/10.24432/C5VP5F}.

\bibitem[{OpenML}(2014)]{OpenMLKin8nm}
{OpenML}.
\newblock kin8nm.
\newblock OpenML data ID 189, version 1, 2014.
\newblock URL \url{https://www.openml.org/d/189}.
\newblock Uploaded 2014-04-23; accessed 2026-07-23.

\bibitem[Bienstock et~al.(2022)Bienstock, Jeong, Shukla, and
  Yun]{Bienstock2022RobustStreamingPCA}
Daniel Bienstock, Minchan Jeong, Apurv Shukla, and Se-Young Yun.
\newblock Robust streaming {PCA}.
\newblock In \emph{Advances in Neural Information Processing Systems},
  volume~35, pages 4231--4243. Curran Associates, Inc., 2022.
\newblock \doi{10.52202/068431-0306}.
\newblock URL
  \url{https://proceedings.neurips.cc/paper_files/paper/2022/hash/1b11d918b08f781a6c194c6c522edfd6-Abstract-Conference.html}.

\end{thebibliography}
\endgroup

\clearpage
\section*{Supplementary material}
\StartSupplement
\section{Proofs and implementation details}\label{supp:proofs}

\subsection{Update rules and numerical conventions}

In the predictive configuration, the update in \mainref{eq:practical_gradient}--\mainref{eq:adam_displacement} uses fixed $\alpha=0.03$, $c=0.5$, $(\beta_1,\beta_2)=(0.9,0.999)$ and $\epsilon=10^{-8}$ outside the square root. The moment step is $1/t$, forgetting is absent, and the optional displacement clip and maximum-basis-change restriction are disabled. The first zero-gradient observation advances the optimizer counter to one. During initial zero moments the initialized span remains unchanged up to QR roundoff; later zero gradients can still produce displacement through nonzero Adam history. The arrays are not transported. The formula box describes the normal finite-data path, not exception handling: the predictive implementation increments the optimizer counter before clipping, whereas the service implementation does so afterward; both use the new counter in bias correction.

Signed QR flips columns only for negative diagonal entries of $R$; zero receives sign $+1$. Rank-deficient inputs are not rejected, and the condition number of $R$ is diagnostic, not an acceptance threshold. Completion at rank deficiency is not a uniquely defined smooth local retraction. Factorization exceptions propagate without rollback.

The predictive core rejects nonfinite raw input before incrementing its sample counter. Nonfinite mapped features instead return an error after advancing that counter but before updating moments or the optimizer, without rollback. A nonfinite post-QR basis triggers Gaussian reinitialization seeded by the current sample counter while retaining advanced moments and optimizer state. These branches do not define accepted-sample skipping. The plain wrapper discards status, but the terminal and rank-aligned studies check every core-update status and reject runs unless all expected updates succeed. Stored clipping counters and norm diagnostics do not supply missing historical branch counts.

Each standardizer transforms before its Welford update. An empty state returns the raw observation; with fewer than two prior observations it centers without scale division. Thereafter variance uses the count as divisor, is clamped below at $10^{-8}$, and a coordinate is divided only when its variance exceeds $10^{-7}$. The input RFF convention in \mainref{eq:implemented_feature_maps} corresponds to $\exp\{-\|x-x'\|_2^2/(2\sigma_x^2)\}$. Regression uses analogous scalar-target features; $D_y=D_x$ in the predictive protocol. The finite-feature bound follows from $\|\phi\|_2^2=(2/D_x)\sum_j\cos^2(\cdot)\le2$. It also holds with changing standardizers, without extending fixed-map probabilistic assumptions.

\paragraph{Gradient derivation and convergence assumptions.}
For $F_t(U)=\tr(U^\top C_tC_t^\top U)$ and a variation $E$, symmetry gives $\mathrm dF_t(U)[E]=2\tr\{(C_tC_t^\top U)^\top E\}$. Differentiating $U^\top U=I_k$ gives $U^\top\Xi+\Xi^\top U=0$; subtracting $U\sym(U^\top G)$ is its orthogonal tangent projection under the Frobenius metric. This proves the direction used before clipping. It does not justify transporting the adaptive arrays, which the implementation does not do, or treating their elementwise-scaled displacement as tangent.

The standard stochastic-retraction result requires smoothness, conditionally unbiased uniformly bounded tangent gradients, Robbins--Monro steps, a twice continuously differentiable retraction and compact containment \citep{bonnabel2013sgd}. Under these conditions objective values converge almost surely and gradient norms vanish. The implemented fixed steps, plug-in moments, ambient adaptation, clipping and evolving standardizers are outside that argument; a separate coupled-recursion analysis would be needed. Neither finite-grid behavior nor exact moment identities supply the missing assumptions.

\subsection{Moment identities, bias, and consistency}

At $t=1$, the recursion gives $\mu_{\phi,1}=\phi_1$, $\mu_{\psi,1}=\psi_1$, $R_1=\phi_1\psi_1^\top$ and $\bar C_1=0$. If a running average is correct at $t-1$, then
\[
(1-t^{-1})\frac1{t-1}\sum_{i<t}a_i+t^{-1}a_t=\frac1t\sum_{i\le t}a_i.
\]
Apply this to both features and their outer product. Expanding the centered products subtracts the mean outer product twice and adds it once, proving \mainref{eq:pathwise_centered_moment} pathwise.

For the probabilistic claims, condition on fixed maps and preprocessing and assume i.i.d. pairs with finite second moments. Write $M_0=\E[\phi\psi^\top]$. Diagonal and off-diagonal terms give
\[
\E[\mu_{\phi,t}\mu_{\psi,t}^\top]
=t^{-2}\{tM_0+t(t-1)\mu_\phi\mu_\psi^\top\}.
\]
Independence is across observations, not within pairs. Subtraction from $\E R_t=M_0$ gives $\E\bar C_t=(t-1)C_*/t$, so $C_t=t\bar C_t/(t-1)$ is unbiased for $t\ge2$ under these conditions only. Cauchy--Schwarz gives $\E|\phi_a\psi_b|\le(\E\phi_a^2\E\psi_b^2)^{1/2}<\infty$. The scalar strong law therefore applies to every raw-moment entry and feature mean. Their limits hold jointly in finite dimensions, and continuity of the mean product proves $\bar C_t\to C_*$ almost surely. Second moments alone supply no concentration rate.

\subsection{Concentration under bounded features}

Fix $t\ge1$ and $\delta\in(0,1)$ under the additional bounded-feature conditions. If $B_xB_y=0$, one feature is zero almost surely and the result is immediate. Otherwise the centered raw outer products and feature increments obey
\[
\begin{aligned}
\|\phi_i\psi_i^\top-M_0\|_F&\le2B_xB_y,\\
\|\phi_i-\mu_\phi\|_2&\le2B_x,&\|\psi_i-\mu_\psi\|_2&\le2B_y.
\end{aligned}
\]
These follow from the outer-product norm identity, Jensen's inequality and the whole-vector bounds. Matrices with the Frobenius inner product form a Hilbert space, as do the feature spaces. For any of these centered sequences $Z_i$ with bound $L$, use the observation filtration and the stopped martingale $S_j=\sum_{i\le j\wedge t}Z_i$. Independence across pairs makes its increments conditionally mean zero, with squared essential-supremum bounds summing to at most $tL^2$. Hilbert-space smoothness constant one in \citet[Theorem~3.5]{pinelis1994} yields
\[
\Pr(\|S_t/t\|\ge u)\le\Pr(\max_{j\le t}\|S_j\|\ge tu)
\le2\exp\{-tu^2/(2L^2)\}.
\]
Set $a_t=\sqrt{2\log(6/\delta)/t}$. Taking $u=La_t$ makes each failure probability at most $\delta/3$. The union bound, requiring no independence between the three events, gives probability at least $1-\delta$ for
\[
\begin{aligned}
\|R_t-M_0\|_F&\le2B_xB_ya_t,\\
\|\mu_{\phi,t}-\mu_\phi\|_2&\le2B_xa_t,\\
\|\mu_{\psi,t}-\mu_\psi\|_2&\le2B_ya_t.
\end{aligned}
\]
The exact centering decomposition is
\[
\begin{aligned}
\mu_{\phi,t}\mu_{\psi,t}^\top-\mu_\phi\mu_\psi^\top
&=(\mu_{\phi,t}-\mu_\phi)\mu_{\psi,t}^\top\\
&\quad+\mu_\phi(\mu_{\psi,t}-\mu_\psi)^\top.
\end{aligned}
\]
The sample target mean in the first term retains the product of mean errors. Since $\|\mu_{\psi,t}\|_2\le B_y$ pathwise and $\|\mu_\phi\|_2\le B_x$, on the same event
\begin{equation}\label{eq:appendix_explicit_concentration}
\begin{aligned}
\opnorm{\bar C_t-C_*}&\le\|\bar C_t-C_*\|_F\\
&\le\|R_t-M_0\|_F+B_y\|\mu_{\phi,t}-\mu_\phi\|_2+B_x\|\mu_{\psi,t}-\mu_\psi\|_2\\
&\le6B_xB_ya_t=6\sqrt2 B_xB_y\sqrt{\log(6/\delta)/t}.
\end{aligned}
\end{equation}
For $t\ge2$, expansion of $C_t-C_*=t(\bar C_t-C_*)/(t-1)+C_*/(t-1)$ also gives
\[
\opnorm{C_t-C_*}\le\frac{t}{t-1}b_t(\delta)+\frac{\opnorm{C_*}}{t-1}.
\]
Stopping at the chosen $t$ proves a fixed-time statement, not simultaneous control over requests or adaptive stopping times.

\subsection{Exact-subspace perturbation and objective-gap bounds}

For two orthonormal $k$-column bases, principal-angle cosines are the singular values of $\widehat U_t^\top U_*$. Idempotence and trace cyclicity give
\begin{equation}\label{eq:appendix_projector_identity}
\|\widehat P_t-P_*\|_F^2=2k-2\|\widehat U_t^\top U_*\|_F^2
=2\sum_{j=1}^k\sin^2\theta_j.
\end{equation}
Under the positive population separator $\Delta_k$, \citet[Theorem~2]{Yu2015DKvariant} bounds the projector distance by
\[
\frac{2\sqrt2}{\Delta_k}\min\{\sqrt{k}\opnorm{M_t-M_*},\|M_t-M_*\|_F\}
\le\frac{2\sqrt{2k}}{\Delta_k}\opnorm{M_t-M_*}.
\]
No positive empirical gap is required for this population-gap bound. With $E_t=\bar C_t-C_*$, expand
\[
M_t-M_*=E_tC_*^\top+C_*E_t^\top+E_tE_t^\top.
\]
Triangle inequality and submultiplicativity prove the second deterministic bound in \mainref{eq:fixed_map_perturbation}. Under all the probabilistic and gap conditions, substitution on the concentration event yields
\[
\|\widehat P_t-P_*\|_F\le\frac{2\sqrt{2k}}{\Delta_k}b_t(\delta)\{2\opnorm{C_*}+b_t(\delta)\}.
\]
This controls the exact empirical basis, not the implemented iterate.

For \mainref{eq:empirical_gap_projector}, express $P$ in an eigenbasis of $M$ and set $s=\sum_{j\le q}(1-P_{jj})=\sum_{j>q}P_{jj}$. The diagonal entries lie in $[0,1]$ and sum to $q$. Its objective loss is therefore at least $(\lambda_q-\lambda_{q+1})s$, while $\|P-P_*^{(q)}\|_F^2=2s$. Rearrangement proves the displayed bound. For binary rank one, projection onto the sole positive direction gives $\rho_1=\cos^2\theta$ and the stated distance identity. For \mainref{eq:within_span_variational}, every contained orthonormal block is $UQ$, $Q=U^\top X\in\St(k,q)$; substitution gives $\tr(Q^\top B_UB_U^\top Q)$ and Ky Fan yields the leading squared-singular-value sum, including zero or tied cutoffs.

For completeness, the classification column identity follows by expanding $\E[(\phi-\mu)(\mathbf1\{y=c\}-p_c)]=p_c(\mu_c-\mu)$. Multiplying the moment by its transpose gives the squared-probability weights. Centered one-hot vectors have zero coordinate sum; multiplying either the population expectation or realized sample-centered sum by $\mathbf1_C$ proves its null vector and the rank upper bound. This argument requires no class-conditional mean for a zero-probability class.

\subsection{Storage and arithmetic complexity}

The core arrays are $R$ of shape $D_x\times D_y$, feature means of lengths $D_x,D_y$, and $U,m,v$ of shape $D_x\times k$, plus scalar counters. Dense maps retain input weights/phases of shapes $D_x\times d_x$ and $D_x$ and, for regression, target weights/phases of shapes $D_y\times d_y$ and $D_y$. Scalers retain means, variances and counts. The predictive implementation additionally allocates feature second-moment buffers of total length $D_x+D_y$ with whitening disabled, internal input-scaler buffers of total length $2d_x$ with internal centering disabled, and a $D_x\times k$ previous-basis diagnostic copy. These unused or diagnostic arrays remain part of that implementation's fixed-dimensional storage; lean numerical harnesses need not allocate them. Scalar/class metadata, monitor lists and random-generator objects are distinct from these named arrays.

The bound excludes raw-data/generator buffers, wrapper objects, retained responses, transient factorization workspaces and process RSS. Dense maps, moment updates, associative gradient products and thin QR supply the main arithmetic terms; conditioning the small QR factor adds $O(k^3)$. None of these bounds establishes observed scaling or latency.

\section{Prediction protocols and paired comparisons}\label{supp:prediction}

\subsection{Data, feature maps, and parameter selection}

Each synthetic final seed $s=100,\ldots,109$ initializes a fresh \texttt{RandomState(s)} and generates 5,000 observations: the first 4,000 train and the last 1,000 test. Seed $10000+s$ permutes training observations only. XOR-Gauss samples equally from centers $(-1,-1),(-1,1),(1,-1),(1,1)$ with labels $0,1,1,0$, adds independent Gaussian noise of SD 0.3, and appends 18 standard-Gaussian nuisance coordinates. Spirals-3 draws independent $u\sim\mathrm{Unif}[0,1]$, $c\sim\mathrm{Unif}\{0,1,2\}$, and $\epsilon\sim N(0,0.2^2)$; its two signal coordinates are
\[
 (r\cos\theta,r\sin\theta),\qquad r=u+\epsilon,\quad
 \theta=4\pi u+2\pi c/3.
\]
The label is $c$, negative radii are retained, and 18 independent Gaussian nuisance coordinates are appended. Friedman-1 uses ten independent uniform $[0,1]$ coordinates and independent $\epsilon\sim N(0,1)$:
\[
y=10\sin(\pi x_1x_2)+20(x_3-0.5)^2+10x_4+5x_5+\epsilon.
\]

HAR preserves the official subject-disjoint test set; development uses \texttt{GroupShuffleSplit}, fraction 0.2 and seed 20260715, yielding 5,588 training and 1,764 validation observations. Sensorless uses the fixed stratified 60/20/20 development construction, then combines development training/validation for the final training partition in \mainref{tab:mt1}. Kin8nm preserves its seed-0 80/20 split and splits only training again at fraction 0.2 with seed 20260715. Real-data seed $s$ changes stream order, input RFF, initialization, and probe random state through $10000+s$, $20000+s$, $30000+s$, and $40000+s$, respectively, rather than held-out membership. Regression target-RFF seed is input-RFF seed plus one.

\begin{table}[!tbp]
\centering
\caption{Predictive settings and development-selected bandwidths. The explicit OKSPCA rate/clip grid and selection rule are distinguished from inherited comparator settings. Bandwidth calibration uses a development-training subset of at most 512 observations; final-test data do not select parameters. Feature budgets appear in \mainref{tab:mt1}.}
\label{tab:st1}
\begingroup\setlength{\tabcolsep}{3pt}\renewcommand{\arraystretch}{1.08}
\begin{tabular*}{\linewidth}{@{\extracolsep{\fill}}p{0.17\linewidth}p{0.76\linewidth}@{}}
\toprule
Item & Setting / selection \\
\midrule
Optimizer & Adam-style QR; rate 0.03; clip 0.5; $\beta_1=0.9$, $\beta_2=0.999$; $\epsilon=10^{-8}$ outside square root; no decay. \\
Development & Rates $\{0.003,0.01,0.03\}$; clips $\{0.5,1.0\}$; XOR-Gauss, HAR, Kin8nm; seeds 0--4; 90 runs. \\
Selection & Minimum average rank; ties by smaller rate, then smaller clip. Both clips tied at rate 0.03. \\
Bandwidth rule & Median positive pairwise distance divided by $\sqrt{2}$ after causal standardization on a fixed development-training subset of at most 512 observations; then frozen. \\
Target mapping & Regression uses the same bandwidth rule on the scalar standardized response; target RFF seed is input RFF seed plus one. \\
Moment & Forgetting factor 1 and empirical step $1/t$; fixed protocol choices. \\
\bottomrule
\end{tabular*}
\par\medskip
\begin{tabular*}{\linewidth}{@{\extracolsep{\fill}}lrr@{}}
\toprule
Task & $\sigma_x$ & $\sigma_y$ (regression) \\
\midrule
XOR-Gauss & 4.49429 & --- \\
Spirals-3 & 4.49160 & --- \\
Friedman-1 & 3.20342 & 0.709610 \\
HAR & 23.0553 & --- \\
Sensorless & 6.40547 & --- \\
Kin8nm & 2.81574 & 0.710298 \\
\bottomrule
\end{tabular*}
\endgroup

\end{table}

\subsection{Comparators and prediction probes}

For $C$-class OVR--RP--CIPLS, each scalar-target reducer receives $+1/-1$ and produces $k$ coordinates. Its compressed output is
\[
r(x)=\Pi^\top\operatorname{col}(z_1(x),\ldots,z_C(x)),\quad
\Pi\in\mathbb R^{Ck\times k},\quad \Pi_{ij}\sim N(0,1/(Ck)).
\]
The projection is sampled once with \texttt{RandomState(20000+s)}. RFF class reducers use identical input maps but store separate copies. Changing $k$ changes both reducer and compression dimensions. Binary and regression CIPLS use scalar targets without this adapter. All CIPLS constructions retain tolerance $10^{-12}$ and no norm clip; CCIPCA's inherited amnesic parameter is two. \Cref{tab:st2} distinguishes these constructions and their unequal tuning opportunities.

\begin{table}[!tbp]
\centering
\caption{Method, target, adapter, and selection identities. Shared input features do not imply shared response construction or symmetric tuning. Raw-X and RFF-X are unreduced controls with different output dimensions from the reduced pipelines.}
\label{tab:st2}
\begingroup\setlength{\tabcolsep}{3pt}\renewcommand{\arraystretch}{1.08}
\begin{tabular*}{\linewidth}{@{\extracolsep{\fill}}>{\raggedright\arraybackslash}p{0.20\linewidth}>{\raggedright\arraybackslash}p{0.15\linewidth}>{\raggedright\arraybackslash}p{0.30\linewidth}>{\raggedright\arraybackslash}p{0.27\linewidth}@{}}
\toprule
Pipeline & Input & Target / construction & Selection opportunity \\
\midrule
OKSPCA & Shared RFF & One-hot / target RFF & Rate and clip selected \\
RFF-CIPLS /\newline RFF--OVR--\newline RP--CIPLS & Shared RFF & Scalar / OVR with fixed $Ck\to k$ RP & Inherited; no method grid \\
CIPLS /\newline OVR--RP--CIPLS & Raw linear & Scalar / OVR with fixed $Ck\to k$ RP & Inherited; no method grid \\
CCIPCA & Raw linear & No target; covariance & Inherited; amnesic 2 \\
RFF-CCIPCA & Shared RFF & No target; covariance & Inherited; no method grid \\
Raw-X & Raw identity & No reduction; $d_x$ outputs & Untuned control \\
RFF-X & Shared RFF & No reduction; $D_x$ outputs & Untuned control \\
\bottomrule
\end{tabular*}
\endgroup

\end{table}

Terminal probes use scikit-learn 1.7.1. Logistic regression uses L2 regularization, \texttt{lbfgs}, tolerance $10^{-4}$, at most 2,000 iterations, no class weights, and an unpenalized intercept. With $n$ training embeddings its objective is mean negative log-likelihood plus $\|W\|_F^2/(2n)$; Ridge uses the dense direct-solver path and minimizes squared-error sum plus $\|w\|_2^2$, also with an unpenalized intercept. Each representation receives its own training-embedding standardizer. The replay definition is \mainref{eq:terminal_probe_pipeline}.

\subsection{Paired effects and unreduced controls}

\begin{samepage}
\Cref{tab:st3} separates three contrasts. The original regression primary contrast is OKSPCA minus RFF-CIPLS; classification rank-cap contrasts use the appropriately named scalar or OVR--RP comparator; the same-target regression interval concerns extracted minus exact. All use 10,000 resamples of ten paired seed differences with replacement, averaging within resample and reporting 2.5th/97.5th percentiles. For the rank-cap and extracted-block extensions, each estimand starts a fresh \texttt{RandomState(50000)} and uses linear percentile interpolation. The original regression inputs and reported intervals are retained, but the exact historical reporting-code and RNG-state binding has not been recovered; the later extensions' seed cannot be assigned to that analysis. Its stored intervals are reproduced as values, not claimed as a bitwise-reproduced bootstrap. Later extensions are conditional descriptions of reused seeds, not independent confirmation. No interval here tests predictive equivalence; exact-minus-full changes in \mainref{tab:mt2} remain separate descriptive quantities.
\par\end{samepage}

The four primary classification caps are one, two, five, and ten for XOR-Gauss, Spirals-3, HAR, and Sensorless. Every reduced pipeline is fitted anew at that cap with its selected or inherited settings unchanged. The classification intervals apply to paired method differences, not to the individual score columns. Regression benchmark dispersion is the sample standard deviation across ten seeds; it is not a confidence interval for a population mean. Reusing a fixed test partition in real-data repetitions also differs from generating a new synthetic sample. Accordingly, the intervals have task-specific conditional interpretations and should not be treated as repeated independent test-set draws.

\begin{table}[!tbp]
\centering
\caption{Retained paired predictive contrasts and their original 95\% percentile-bootstrap intervals. Classification contrasts are rank-cap OKSPCA minus the named comparator; primary regression contrasts are OKSPCA minus RFF-CIPLS; same-target regression contrasts are extracted minus exact. The last are not exact-minus-full intervals.}
\label{tab:st3}
\begingroup\setlength{\tabcolsep}{3pt}\renewcommand{\arraystretch}{1.08}
\begin{tabular*}{\linewidth}{@{\extracolsep{\fill}}llr@{}}
\toprule
Task & Metric & OKSPCA $-$ RFF-CIPLS-based (95\% CI) \\
\midrule
XOR-Gauss & Macro-F1 & -0.0002 [-0.0010, +0.0008] \\
Spirals-3 & Macro-F1 & +0.0170 [+0.0029, +0.0343] \\
HAR & Macro-F1 & +0.0504 [+0.0334, +0.0667] \\
Sensorless & Macro-F1 & +0.1513 [+0.1186, +0.1860] \\
Friedman-1 & $R^2$ & -0.0564 [-0.0706, -0.0416] \\
Kin8nm & $R^2$ & -0.0416 [-0.0489, -0.0352] \\
\bottomrule
\end{tabular*}
\par\medskip
\begin{tabular*}{\linewidth}{@{\extracolsep{\fill}}llr@{}}
\toprule
Task & Metric & Extracted $-$ exact (95\% CI) \\
\midrule
Friedman-1 & $R^2$ & -0.00014 [-0.00153, +0.00099] \\
Kin8nm & $R^2$ & -0.00003 [-0.00021, +0.00013] \\
\bottomrule
\end{tabular*}
\endgroup

\end{table}

\begin{table}[!tbp]
\centering
\caption{No-reduction controls and their original-budget OKSPCA reference. Entries are ten-final-seed means and sample standard deviations. Raw-X has $d_x$ coordinates and RFF-X has $D_x$; OKSPCA uses the displayed original $k_{\rm old}$. Classification entries are not rank-cap control experiments. Different output dimensions preclude including these controls in reduced-method ranks.}
\label{tab:st4}
\begingroup\setlength{\tabcolsep}{3pt}\renewcommand{\arraystretch}{1.08}
\begin{tabular*}{\linewidth}{@{\extracolsep{\fill}}llrrrr@{}}
\toprule
Task & Metric & $k_{\rm old}$ & Old OKSPCA & Raw-X & RFF-X \\
\midrule
XOR-Gauss & Macro-F1 & 4 & 0.605 $\pm$ 0.035 & 0.504 $\pm$ 0.016 & 0.660 $\pm$ 0.032 \\
Spirals-3 & Macro-F1 & 5 & 0.369 $\pm$ 0.013 & 0.378 $\pm$ 0.009 & 0.360 $\pm$ 0.010 \\
Friedman-1 & $R^2$ & 4 & 0.663 $\pm$ 0.042 & 0.723 $\pm$ 0.016 & 0.853 $\pm$ 0.015 \\
HAR & Macro-F1 & 20 & 0.883 $\pm$ 0.009 & 0.954 $\pm$ 0.001 & 0.939 $\pm$ 0.005 \\
Sensorless & Macro-F1 & 12 & 0.773 $\pm$ 0.025 & 0.913 $\pm$ 0.000 & 0.945 $\pm$ 0.003 \\
Kin8nm & $R^2$ & 4 & 0.413 $\pm$ 0.025 & 0.404 $\pm$ 0.000 & 0.657 $\pm$ 0.033 \\
\bottomrule
\end{tabular*}
\endgroup

\end{table}
\FloatBarrier

\section{Same-target diagnostic protocols}\label{supp:diagnostics}

\subsection{Primary-rank classification reconstruction}

The 40 fits in \mainref{tab:mt3} start at the primary classification ranks with the original inputs and settings, reproduce the terminal scores, and retain five checkpoints each. They reconstruct states rather than recover historical states. Numerical rank counts squared singular values strictly above $10^{-10}\sigma_1^2$ in the full saved moment; comparison rank is the smaller of this rank and the trained cap. At equal rank, geometry uses the maintained span; at lower positive rank it uses within-span extraction without changing the state. The original-rank terminal probe is retained. A smaller exact comparison dimension remains a mixed-dimension comparison, without adding a third extracted-block probe.

At rank zero, basis and normalized geometric diagnostics are undefined and the exact objective is zero. For positive rank with $0<J^*\le10^{-15}$, raw objectives remain recorded but normalized diagnostics are undefined. An exact tie has zero boundary gap; the numerical-tie flag uses gap at most $10^{-10}\lambda_1$. These are reconstruction conventions, distinct from the service rule in \mainref{sec:services}. \Cref{tab:st5} retains all five Sensorless seed-104 checkpoints, including the adverse midpoint. They do not establish the deviation's duration or cause.

For two orthonormal equal-rank bases, principal angles are determined by the singular values of their cross-product; the largest acute angle reports the least-aligned direction. Projector distance is $\|UU^\top-VV^\top\|_F$, without division by rank. Objective ratios use the same empirical moment and comparison dimension. These statistics cannot be interchanged with prediction scores, and a tied boundary does not select a unique projector even when the leading objective value remains well defined.

\begin{table}[!tbp]
\centering
\caption{The five retained Sensorless seed-104 checkpoints of the directly trained primary-rank reconstruction. Rank, objective capture, projector distance, and maximum angle retain the same-state definitions. Geometric entries are rounded to six decimals; displayed $\rho=1.000000$ does not assert exact equality. The observation-23,404 deviation is a measured checkpoint, not a uniform pathwise statement.}
\label{tab:st5}
\begingroup\setlength{\tabcolsep}{3pt}\renewcommand{\arraystretch}{1.08}
\begin{tabular*}{\linewidth}{@{\extracolsep{\fill}}lrrrrrr@{}}
\toprule
Observation & $k_{\rm train}$ & $r_{\tau}$ & $q$ & $\rho$ & $\theta_{\max}$ (deg) & $d_P$ \\
\midrule
4681 & 10 & 10 & 10 & 0.999978 & 2.622743 & 0.087174 \\
11702 & 10 & 10 & 10 & 0.999922 & 0.689740 & 0.029430 \\
23404 & 10 & 10 & 10 & 0.970687 & 78.099077 & 1.898569 \\
35106 & 10 & 10 & 10 & 0.999905 & 0.655705 & 0.023416 \\
46807 & 10 & 10 & 10 & 1.000000 & 0.163814 & 0.006250 \\
\bottomrule
\end{tabular*}
\endgroup

\end{table}

\subsection{Regression representation comparisons}

\begin{samepage}
The regression endpoints use final seeds 100--109 from the larger-span terminal study. Full is the maintained $k=4$ representation; extracted is its maximizing within-span block as in \mainref{eq:within_span_variational}; exact is the leading left singular block of the same terminal $\bar C_T$. Positive rescaling $C_T=T\bar C_T/(T-1)$ does not change its singular subspaces. The full-moment threshold is $10^{-10}\sigma_1^2$, with $q_T=\min(k,r_\tau)$; both regressions have $q_T=k=4$. Each representation has its own standardizer and penalized probe.
\par\end{samepage}

Extraction can rotate coordinates even at equal rank. Mean full-minus-extracted $R^2$ is $-3.897\times10^{-7}$ for Kin8nm and $-5.315\times10^{-7}$ for Friedman-1; maximum absolute seed differences are $1.358\times10^{-6}$ and $1.956\times10^{-6}$. Their small size does not establish rotation invariance. \Cref{tab:st6} reports extracted-to-exact geometry. The extracted-minus-exact intervals in \cref{tab:st3} and descriptive exact-minus-full changes in \mainref{tab:mt2} compare different paired quantities.

\begin{table}[!htbp]
\centering
\caption{Regression terminal extracted-to-exact objective and geometry, averaged over ten final seeds at $q_T=4$. Projector distance is unnormalized Frobenius distance; angle is the maximum principal angle in degrees. These are the larger-span study's regression endpoints, not direct classification-rank reconstructions.}
\label{tab:st6}
\begingroup\setlength{\tabcolsep}{3pt}\renewcommand{\arraystretch}{1.08}
\begin{tabular*}{\linewidth}{@{\extracolsep{\fill}}lrrrr@{}}
\toprule
Task & $q_T$ & Objective ratio $\rho$ & $d_P$ & $\theta_{\max}$ \\
\midrule
Friedman-1 & 4 & 0.999969 & 0.1426 & 5.347$^\circ$ \\
Kin8nm & 4 & 0.999980 & 0.0300 & 1.016$^\circ$ \\
\bottomrule
\end{tabular*}
\endgroup

\end{table}
\FloatBarrier

\section{Bounded population-recovery protocol}\label{supp:population}

This experiment uses identity features $\phi(x)=x$ and $\psi(y)=y$, with no RFF or raw-data standardizer. Its independently specified tracker follows the moment and maintenance stages B--C of \mainref{alg:okspca_practical}: empirical step $1/t$, Bessel-corrected gradient moment for $t>1$ (zero at $t=1$), ambient Adam with fixed $\alpha=0.03$, $(\beta_1,\beta_2)=(0.9,0.999)$ and $\epsilon=10^{-8}$ outside the square root. The tangent gradient is clipped globally at Frobenius norm $0.5$, using the over-threshold denominator stabilizer $10^{-12}$. There is no learning-rate decay, adaptive scheduling or moment transport. Adam arrays and its counter start at zero; each sample advances the counter and applies signed reduced QR. Initialization is specified below. The RFF/scaler stages A and D of the predictive algorithm are absent here.

Draw independent standard Gaussian $g_x\in\mathbb R^{32}$ and $g_e\in\mathbb R^8$, and define
\[
x=\sqrt{32}\,g_x/\|g_x\|_2,\qquad
e=\sqrt8\,g_e/\|g_e\|_2,\qquad y=C_*^\top x+Be,
\]
where $B=(I-C_*^\top C_*)^{1/2}$ is the symmetric positive-semidefinite square root. The four positive squared singular values of $C_*$ are $(0.49,0.36,0.25,0.15)$, all others zero. Thus $\|C_*\|_{\rm op}=0.7<1$, the root exists, and $(D_x,D_y,r,k)=(32,8,4,3)$; the fourth direction defines the selected boundary gap.

Spherical symmetry and independence give centered coordinates, $\operatorname{Cov}(x)=I_{32}$, $\operatorname{Cov}(e)=I_8$, and
\[
\operatorname{Cov}(x,y)=C_*,\qquad
\operatorname{Cov}(y)=C_*^\top C_*+BB^\top=I_8.
\]
Moreover, $\|x\|_2=\sqrt{32}$, $\|e\|_2=\sqrt8$, and $\|y\|_2\le\|C_*\|_{\rm op}\sqrt{32}+\|B\|_{\rm op}\sqrt8$ almost surely. These fixed-feature, bounded coordinates satisfy the stated population construction without time-varying standardization.

Signed reduced QR of counter-generated Gaussian matrices with orientation seeds 81001 and 81002 fixes the left/right singular vectors across all runs. Seeds 0--19 vary the spherical streams and separately namespaced Gaussian three-column tracker initialization, not population orientations. Sample sizes $256,512,\ldots,32{,}768$ are nested within seed. The objective curves in \mainref{fig:mf2} are $\rho_3(U_{\rm online};\bar C_n)$, $\rho_3(U_{\rm online};C_*)$, and $\rho_3(U_{\rm exact};C_*)$; the empirical-exact ratio is one by construction. Projector pairs are defined in \mainref{eq:population_distances}. Pointwise seed medians and interquartile ranges describe this finite grid, not independent checkpoints or an estimated convergence rate.

\section{Numerical-service protocols and comparisons}\label{supp:services}

\subsection{Streams, query schedules, and parameter selection}

For the workloads in \mainref{tab:mt4}, draw $x_i\sim N(0,I_{d_x})$. Classification uses $C\in\{2,4,11\}$ and selects the label maximizing
\[
1.25\{x_{i1}\cos(2\pi c/C)+x_{i2}\sin(2\pi c/C)\}+g_{ic},
\quad c=0,\ldots,C-1,
\]
with independent standard Gumbel $g_{ic}$ and the lowest class index breaking exact ties. Regression uses independent $\epsilon_i\sim N(0,1)$ and
\[
y_i=\sin(x_{i1})+0.5(x_{i2}^2-1)+0.75x_{i3}x_{i4}+0.25x_{i5}+0.25\epsilon_i.
\]
Remaining raw coordinates are Gaussian nuisances. All methods within regime and seed share stream, maps, order, and the signed-QR Gaussian initial basis. Input RFF bandwidth is $\sqrt{d_x}$; regression target bandwidth is one in causally standardized units. Each service computes the frontend and moment itself. These computational generators have no predictive endpoint.

The Adam service uses fixed base rate $0.03$, $\beta_1=0.9$, $\beta_2=0.999$, global Frobenius gradient clip $0.5$, and $\epsilon=10^{-8}$ outside the square root. The clipping denominator uses $10^{-12}$. These settings belong to the service implementation; its final runs do not retune them against PG's development-selected step.

For the structured multiclass service, Helmert column $j=1,\ldots,C-1$ equals $1/\sqrt{j(j+1)}$ in rows $1,\ldots,j$, $-j/\sqrt{j(j+1)}$ in row $j+1$, and zero elsewhere. Hence $H_CH_C^\top=I-\mathbf1\mathbf1^\top/C$. Reduced QR of $\bar C_tH_C$ followed by SVD of the triangular factor determines the singular values and input directions; rank is not inferred from QR diagonal entries. Plain QR suffices for an exact full contrast span only if $D_x\ge C-1$, $\rank(\bar C_t)=C-1$, and the requested rank is $C-1$. Deficient rank, smaller requested blocks, or canonical singular coordinates require the stated spectral work. Floating-point residual checks against the original moment remain outside timing.

Final seeds are 6200--6204; PG development seeds are 6100--6102. Periodic requests use positive multiples of 1, 10, 100, or 1000 and termination, with duplicates removed. Bernoulli requests use probability 0.01 with the fixed seed-index hash construction and include termination. Cache updates retain periods ten or one hundred, with no extra terminal refresh. Same-timestamp extraction can reuse factors; cache age is measured against the request index.

Returned bases carry their actual dimension and update timestamp. Approximate services use the spectrum of their inspected small matrix to determine output rank; caches retain the rank at their last scheduled factorization. A no-target response describes the inspected matrix, whereas a zero full-reference rank describes the offline target. Thus these two events need not coincide for approximate or stale services, and request status cannot be reconstructed from a conditional objective mean alone.

For PG, each $\gamma\in\{0.1,0.3,1,3\}$ is tested on the three development seeds. Let $\mathcal T_s$ contain positive-reference observations and define
\begin{equation}\label{eq:pg_selection_loss}
\ell_{s,t}(\gamma)=
\begin{cases}1-\rho_{s,t}(\gamma),&\text{valid same-rank output},\\
1,&\text{missing or unequal-rank output},\end{cases}
\quad
L(\gamma)=\frac13\sum_{s=1}^3\frac{\sum_{t\in\mathcal T_s}\ell_{s,t}(\gamma)}{|\mathcal T_s|}.
\end{equation}
Selection first minimizes total shortfalls, then chooses the smallest $\gamma$ within $10^{-12}$ of the minimum loss among those candidates. Zero-reference observations are excluded from loss; numerical failures stop the trajectory. The unit penalty is a development criterion, not an imputation for final undefined geometry. \Cref{tab:st7} retains all 20 candidates, including R256's selected $\gamma=0.1$.

\begin{table}[!tbp]
\centering
\caption{All PG development candidates. Cells give total shortfalls and equally seed-weighted prescribed loss; the selected candidate follows shortfall, loss, and step-size priority. The development grid does not constitute a matched tuning budget for fixed-setting Adam.}
\label{tab:st7}
\begingroup\setlength{\tabcolsep}{3pt}\renewcommand{\arraystretch}{1.08}
\begin{tabular*}{\linewidth}{@{\extracolsep{\fill}}lrrrr@{}}
\toprule
Regime & $\gamma=0.1$ & $\gamma=0.3$ & $\gamma=1$ & $\gamma=3$ \\
\midrule
C2 & 0 / 0.00955 & 0 / 0.00358 & \textbf{0 / 0.00076} & 0 / 0.16788 \\
C4 & 0 / 0.03017 & 0 / 0.01075 & 0 / 0.00239 & \textbf{0 / 0.00081} \\
C11 & 0 / 0.09772 & 0 / 0.03602 & 0 / 0.00981 & \textbf{0 / 0.00292} \\
R128 & 0 / 0.03005 & 0 / 0.01138 & \textbf{0 / 0.00304} & 0 / 0.09884 \\
R256 & \textbf{10 / 0.07625} & 20 / 0.03354 & 18 / 0.01298 & 16 / 0.07763 \\
\bottomrule
\end{tabular*}
\endgroup

\end{table}

\subsection{Timing protocol and repetitions}

\begin{samepage}
The recorded execution environment was Windows on an AMD Ryzen 5 7500F six-core processor, using Python 3.13.6, NumPy 2.3.2 and SciPy 1.16.1. The loaded NumPy and SciPy OpenBLAS libraries were versions 0.3.30 and 0.3.28, respectively, each reporting one thread; the process affinity mask was one. Scientific arrays used float64. Generic exact and small-matrix factorizations called \texttt{numpy.linalg.svd} with \texttt{full\_matrices=False}, then selected the required columns. No explicit LAPACK-driver option was supplied. The retained environment record does not identify the Windows edition/build or record a per-call LAPACK-driver trace; these are not inferred from the present machine.
\par\end{samepage}

\begin{samepage}
A0 is measured in a fresh process after initialization. It includes causal preprocessing, features, moments, maintenance, and required output extraction, signed QR, copying, and arithmetic/shape/finiteness checks. Exact-each pays factorization during maintenance. Libraries use one thread and the process one logical processor. Stream preparation, offline-reference construction, quality comparisons, hashing, and serialization are excluded. \Cref{tab:st8} retains the 125 paired base timings through their 25 group summaries, including initialization-inclusive results and the anomalous R128 observation.
\par\end{samepage}

\begin{table}[!tbp]
\centering
\caption{All 25 Adam/exact-request schedule groups. Paired A0 medians, ranges, and Adam-faster counts retain all five base seed pairs per group. Initialization-inclusive ratios are separate from primary service time. No anomalous timing is removed.}
\label{tab:st8}
\begingroup\setlength{\tabcolsep}{3pt}\renewcommand{\arraystretch}{1.08}
\begin{tabular*}{\linewidth}{@{\extracolsep{\fill}}llrrrrr@{}}
\toprule
Regime & Spacing & Median & Min & Max & \shortstack{Faster\\seeds / 5} & \shortstack{Init.-included\\median ratio} \\
\midrule
C2 & 1 & 1.726 & 1.617 & 1.841 & 0 & 1.726 \\
C2 & 10 & 2.210 & 2.013 & 2.427 & 0 & 2.209 \\
C2 & 100 & 2.242 & 2.090 & 2.360 & 0 & 2.241 \\
C2 & 1000 & 2.044 & 1.846 & 2.424 & 0 & 2.044 \\
C2 & Bernoulli $0.01$ & 2.199 & 1.804 & 2.289 & 0 & 2.198 \\
C4 & 1 & 1.806 & 1.716 & 1.911 & 0 & 1.806 \\
C4 & 10 & 2.259 & 2.232 & 2.759 & 0 & 2.258 \\
C4 & 100 & 2.488 & 2.337 & 2.584 & 0 & 2.486 \\
C4 & 1000 & 2.452 & 2.117 & 2.648 & 0 & 2.452 \\
C4 & Bernoulli $0.01$ & 2.541 & 2.246 & 2.758 & 0 & 2.540 \\
C11 & 1 & 1.513 & 1.251 & 2.013 & 0 & 1.512 \\
C11 & 10 & 2.628 & 2.317 & 2.887 & 0 & 2.627 \\
C11 & 100 & 2.830 & 2.615 & 2.873 & 0 & 2.829 \\
C11 & 1000 & 2.886 & 2.750 & 3.239 & 0 & 2.884 \\
C11 & Bernoulli $0.01$ & 2.899 & 2.654 & 3.118 & 0 & 2.890 \\
R128 & 1 & 0.277 & 0.261 & 0.286 & 5 & 0.277 \\
R128 & 10 & 0.923 & 0.892 & 1.144 & 3 & 0.923 \\
R128 & 100 & 1.841 & 1.744 & 1.964 & 0 & 1.841 \\
R128 & 1000 & 2.097 & 0.287 & 2.613 & 1 & 2.096 \\
R128 & Bernoulli $0.01$ & 1.719 & 1.637 & 1.775 & 0 & 1.719 \\
R256 & 1 & 0.085 & 0.081 & 0.101 & 5 & 0.085 \\
R256 & 10 & 0.567 & 0.497 & 0.932 & 5 & 0.567 \\
R256 & 100 & 1.625 & 1.332 & 2.283 & 0 & 1.624 \\
R256 & 1000 & 1.622 & 1.455 & 2.021 & 0 & 1.622 \\
R256 & Bernoulli $0.01$ & 1.435 & 1.130 & 2.321 & 0 & 1.434 \\
\bottomrule
\end{tabular*}
\endgroup

\end{table}

\begin{samepage}
Pass B measures inclusive 64-observation batch means and synchronous query durations nested within batches. Its quantiles retain all completed calls, including unavailable/no-target responses, without subtracting overhead or dropping resolution-limited timings. These are not individual-update tail latencies, queueing delays, or additive components of A0. Untimed pass C measures quality and state correspondence. Retaining responses may grow harness memory even though named numerical arrays have fixed dimensions.
\par\end{samepage}

\begin{table}[!tbp]
\centering
\caption{All prescribed C4 seed-6200, period-ten repeats: three Adam, three exact-request, and six accumulation-only measurements. Wall and CPU times are in milliseconds, with initialization-inclusive values separate. Accumulation A/B identify repeated launches, not different algorithms. These 12 measurements add no input seeds and establish no grid-wide timing stability.}
\label{tab:st9}
\begingroup\setlength{\tabcolsep}{3pt}\renewcommand{\arraystretch}{1.08}
\begin{tabular*}{\linewidth}{@{\extracolsep{\fill}}lrrrrr@{}}
\toprule
Service / control & Repeat & \shortstack{Service wall\\(ms)} & \shortstack{Service CPU\\(ms)} & \shortstack{Total wall\\(ms)} & \shortstack{Total CPU\\(ms)} \\
\midrule
Adam & 0 & 683.60 & 609.38 & 683.79 & 609.38 \\
Adam & 1 & 692.76 & 609.38 & 692.97 & 609.38 \\
Adam & 2 & 661.37 & 656.25 & 661.61 & 656.25 \\
Exact-request & 0 & 302.66 & 296.88 & 302.85 & 296.88 \\
Exact-request & 1 & 295.91 & 265.62 & 296.09 & 265.62 \\
Exact-request & 2 & 299.18 & 296.88 & 299.54 & 296.88 \\
Accum.-A & 0 & 247.13 & 203.12 & 247.31 & 203.12 \\
Accum.-A & 1 & 260.19 & 250.00 & 260.37 & 250.00 \\
Accum.-A & 2 & 266.83 & 218.75 & 267.03 & 218.75 \\
Accum.-B & 0 & 249.48 & 218.75 & 249.70 & 218.75 \\
Accum.-B & 1 & 233.59 & 234.38 & 233.80 & 234.38 \\
Accum.-B & 2 & 258.21 & 218.75 & 258.42 & 218.75 \\
\bottomrule
\end{tabular*}
\endgroup

\end{table}

\subsection{Quality and request coverage}

\begin{samepage}
\Cref{tab:st10,tab:st11} cover all 33 applicable every-observation combinations: six services in every regime plus binary C2 and contrast C4/C11. Time ratios first pair A0 by seed against exact-request, then take the median. Absolute medians come directly from measured durations. Quality first averages valid positive-reference, matched-rank requests within seed, then averages five seed means equally. Counts retain request denominators; stale overlaps validity and rank categories. This grid does not replace the distinct schedule comparison. Structured services have near-unit objective ratios with measured floating-point residuals; exact-each and exact-request differ in when they pay for the same current target.
\par\end{samepage}

\begin{table}[!tbp]
\centering
\caption{Complete classification every-observation comparison: all 21 applicable method--regime combinations. Cost, conditional geometry, matched/positive-reference counts, shortfalls, and overlapping stale counts are read jointly. Conditional means do not measure whole-grid success rates.}
\label{tab:st10}
\begingroup\setlength{\tabcolsep}{3pt}\renewcommand{\arraystretch}{1.08}
\begin{tabular*}{\linewidth}{@{\extracolsep{\fill}}lrrrrrrr@{}}
\toprule
Service & \shortstack{Paired A0\\ratio} & Mean $\rho$ & Mean $d_P$ & \shortstack{Mean angle\\(deg)} & \shortstack{Matched /\\positive} & Shortfall & Stale \\
\midrule
\multicolumn{8}{l}{\textit{C2: Exact-request A0 wall median = 492.9 ms}} \\
\shortstack[l]{Exact-\\request} & 1.000 & 1.0000 & 0 & 6.6e-07 & 20473/20473 & 0 & 0 \\
Adam & 1.726 & 0.9946 & 0.0383 & 1.59 & 20473/20473 & 0 & 0 \\
PG & 1.798 & 0.9994 & 0.0013 & 0.07 & 20473/20473 & 0 & 0 \\
Cache-10 & 0.714 & 0.9977 & 0.0271 & 1.11 & 20435/20473 & 38 & 18390 \\
Cache-100 & 0.572 & 0.9920 & 0.0731 & 2.99 & 19985/20473 & 488 & 19785 \\
Exact-each & 0.984 & 1.0000 & 0 & 6.6e-07 & 20473/20473 & 0 & 0 \\
Binary & 0.878 & 1.0000 & 1.6e-14 & 4.5e-07 & 20473/20473 & 0 & 0 \\
\addlinespace[4pt]
\multicolumn{8}{l}{\textit{C4: Exact-request A0 wall median = 594.2 ms}} \\
\shortstack[l]{Exact-\\request} & 1.000 & 1.0000 & 0 & 1.4e-06 & 20474/20474 & 0 & 0 \\
Adam & 1.806 & 0.9898 & 0.199 & 7.33 & 20474/20474 & 0 & 0 \\
PG & 1.752 & 0.9990 & 0.0472 & 1.91 & 20474/20474 & 0 & 0 \\
Cache-10 & 0.624 & 0.9962 & 0.102 & 3.56 & 20427/20474 & 47 & 18390 \\
Cache-100 & 0.569 & 0.9797 & 0.308 & 10.36 & 19985/20474 & 489 & 19785 \\
Exact-each & 1.019 & 1.0000 & 0 & 1.4e-06 & 20474/20474 & 0 & 0 \\
Contrast & 1.129 & 1.0000 & 1.5e-15 & 1.3e-06 & 20474/20474 & 0 & 0 \\
\addlinespace[4pt]
\multicolumn{8}{l}{\textit{C11: Exact-request A0 wall median = 880.8 ms}} \\
\shortstack[l]{Exact-\\request} & 1.000 & 1.0000 & 0 & 2.2e-06 & 20473/20473 & 0 & 0 \\
Adam & 1.513 & 0.9812 & 0.653 & 16.49 & 20473/20473 & 0 & 0 \\
PG & 1.458 & 0.9971 & 0.243 & 6.89 & 20473/20473 & 0 & 0 \\
Cache-10 & 0.516 & 0.9966 & 0.229 & 6.26 & 20363/20473 & 110 & 18390 \\
Cache-100 & 0.423 & 0.9756 & 0.727 & 15.04 & 19985/20473 & 488 & 19785 \\
Exact-each & 1.022 & 1.0000 & 0 & 2.2e-06 & 20473/20473 & 0 & 0 \\
Contrast & 0.948 & 1.0000 & 3.8e-15 & 1.7e-06 & 20473/20473 & 0 & 0 \\
\bottomrule
\end{tabular*}
\endgroup

\end{table}

\begin{table}[!tbp]
\centering
\caption{Complete regression every-observation comparison: all 12 applicable combinations, including exact-each. Statistics have the same seed weighting and request conditioning as \cref{tab:st10}. R128 and R256 are distinct workloads, not replicates of a width intervention.}
\label{tab:st11}
\begingroup\setlength{\tabcolsep}{3pt}\renewcommand{\arraystretch}{1.08}
\begin{tabular*}{\linewidth}{@{\extracolsep{\fill}}lrrrrrrr@{}}
\toprule
Service & \shortstack{Paired A0\\ratio} & Mean $\rho$ & Mean $d_P$ & \shortstack{Mean angle\\(deg)} & \shortstack{Matched /\\positive} & Shortfall & Stale \\
\midrule
\multicolumn{8}{l}{\textit{R128: Exact-request A0 wall median = 5,642 ms}} \\
\shortstack[l]{Exact-\\request} & 1.000 & 1.0000 & 0 & 1.8e-06 & 20475/20475 & 0 & 0 \\
Adam & 0.277 & 0.9905 & 0.658 & 28.83 & 20475/20475 & 0 & 0 \\
PG & 0.280 & 0.9971 & 0.447 & 17.11 & 20475/20475 & 0 & 0 \\
Cache-10 & 0.193 & 0.9979 & 0.0951 & 3.48 & 20435/20475 & 40 & 18390 \\
Cache-100 & 0.107 & 0.9930 & 0.299 & 9.92 & 19985/20475 & 490 & 19785 \\
Exact-each & 0.979 & 1.0000 & 0 & 1.8e-06 & 20475/20475 & 0 & 0 \\
\addlinespace[4pt]
\multicolumn{8}{l}{\textit{R256: Exact-request A0 wall median = 18,840 ms}} \\
\shortstack[l]{Exact-\\request} & 1.000 & 1.0000 & 0 & 2.2e-06 & 10235/10235 & 0 & 0 \\
Adam & 0.085 & 0.9719 & 2.31 & 87.67 & 10210/10235 & 25 & 0 \\
PG & 0.113 & 0.9201 & 2.98 & 87.94 & 10226/10235 & 9 & 0 \\
Cache-10 & 0.134 & 0.9931 & 0.252 & 8.67 & 10187/10235 & 48 & 9175 \\
Cache-100 & 0.047 & 0.9612 & 0.815 & 23.28 & 9745/10235 & 490 & 9645 \\
Exact-each & 1.016 & 1.0000 & 0 & 2.2e-06 & 10235/10235 & 0 & 0 \\
\bottomrule
\end{tabular*}
\endgroup

\end{table}

\begin{figure}[!tbp]
\centering
\includegraphics[width=\textwidth]{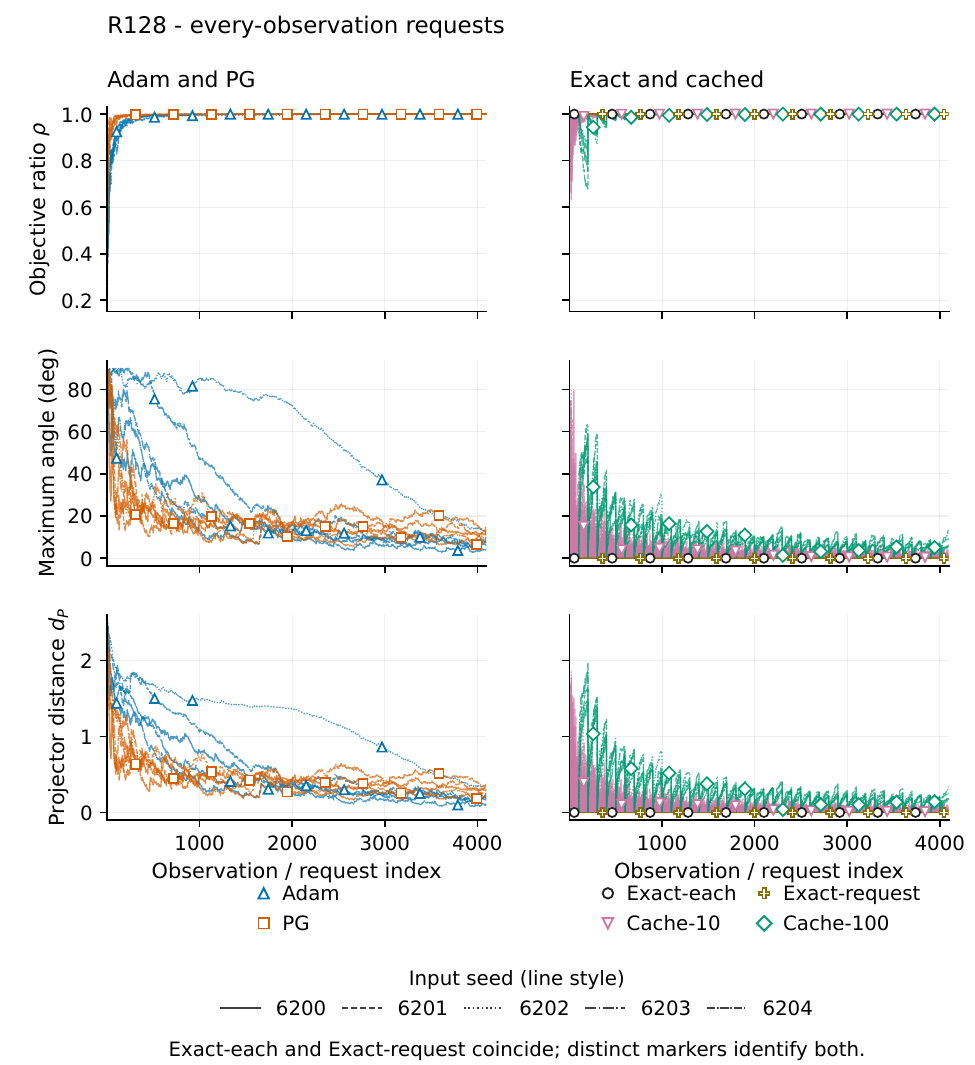}
\caption{Complete every-observation R128 trajectories for all five seeds, including early behavior and undefined segments. Objective, maximum-angle, and projector-distance rows use comparable axes across Adam/PG and exact/cached columns. Conditional cache quality does not remove early shortfalls or stale returns. Line styles identify seeds (6202 is dotted); colours and sparse markers identify services.}
\label{fig:sf1}
\end{figure}

Across all schedules, valid-basis, uninitialized-cache, and no-identified-target responses number 685,913, 2,975, and 140, respectively. Their sum is 689,028; these status counts differ from the positive-reference coverage partition in \mainref{sec:services}. Current-rank increases can make an initialized cache rank insufficient. Neither missing nor unequal-rank geometry is filled with zero.

\subsection{R256 delivered-basis correspondence}

\begin{samepage}
\Cref{tab:st12} compares the five terminal delivered query-cache bases with their saved exact references, not with raw maintained bases. Their maximum angles range from $84.109^\circ$ to $89.407^\circ$. The separate 5,120-request, latter-half description uses recorded scalar diagnostics and a post hoc threshold; it is not a recomputation from all corresponding saved states. For the five endpoints, $d_P^2/2=\sum_j\sin^2\theta_j$ is approximately $1.68,1.94,3.00,2.95,2.21$ in seed order. Every sum exceeds one, excluding a single misaligned direction with all others exactly aligned, but not implying that all directions are orthogonal.
\par\end{samepage}

\begin{table}[!tbp]
\centering
\caption{All five R256 Adam terminal responses. Full-reference rank, capped comparison rank, and actual output rank are distinct. The relative boundary gap is the stored gap divided by the capped exact objective; a positive absolute gap need not be large relatively. State correspondence applies to these five delivered responses.}
\label{tab:st12}
\begingroup\setlength{\tabcolsep}{3pt}\renewcommand{\arraystretch}{1.08}
\begin{tabular*}{\linewidth}{@{\extracolsep{\fill}}lrrrrrrr@{}}
\toprule
Seed & $r_{\rm ref}$ & $q_{\rm ref}$ & $q_{\rm out}$ & $\rho$ & $\theta_{\max}$ (deg) & $d_P$ & Gap/$J^*$ \\
\midrule
6200 & 16 & 8 & 8 & 0.996425 & 84.109 & 1.8307 & 0.000446 \\
6201 & 15 & 8 & 8 & 0.995430 & 88.699 & 1.9698 & 0.000497 \\
6202 & 13 & 8 & 8 & 0.994618 & 86.733 & 2.4510 & 0.000192 \\
6203 & 14 & 8 & 8 & 0.992189 & 87.106 & 2.4293 & 0.000382 \\
6204 & 14 & 8 & 8 & 0.995622 & 89.407 & 2.1019 & 0.000342 \\
\bottomrule
\end{tabular*}
\endgroup

\end{table}
\FloatBarrier

\end{document}